\documentclass{article} 
\usepackage{iclr2017_conference,times}
\usepackage{hyperref}
\usepackage{url}
\usepackage{color}
\usepackage{graphicx}
\usepackage{morefloats}
\usepackage{subcaption}
\usepackage{caption}
\usepackage[normalem]{ulem}
\usepackage{adjustbox}
\usepackage{fancybox}
\usepackage{array}
\usepackage{booktabs}
\usepackage{multirow}
\usepackage{array}
\usepackage{xspace}
\usepackage{cleveref}
\usepackage{longtable}

\newif\ifshort
\shortfalse

\newif\ifarxiv
\arxivtrue

\title{Delving into Transferable Adversarial Examples and Black-box Attacks}

\author{Yanpei Liu$^*$, Xinyun Chen\thanks{Work is done while visiting UC Berkeley.}\\
Shanghai Jiao Tong University\\
\And
Chang Liu, Dawn Song\\
University of the California, Berkeley \\
}

\newcommand{\clarifai}{Clarifai.com\xspace}

\newcommand{\chang}[1]{{\color{blue}[Chang: #1]}}
\newcommand{\changcomment}[1]{}

\newcommand{\xinyun}[1]{{\color{blue}[Xinyun: #1]}}
\newcommand{\eat}[1]{}
\newcolumntype{C}[1]{>{\centering\let\newline\\\arraybackslash\hspace{0pt}}m{#1}}

\begin{document}

\iclrfinalcopy

\maketitle

\begin{abstract}
An intriguing property of deep neural networks is the existence
of adversarial examples, which can transfer among different architectures.
These transferable adversarial examples may severely hinder
deep neural network-based applications. Previous works mostly study
the transferability using small scale datasets. In this work, we are the first to conduct an extensive study of the transferability over large models and a large scale
dataset, and we are also the first to study the transferability of targeted adversarial examples with their target labels. We study both \emph{non-targeted} and \emph{targeted} adversarial examples, and show that while transferable non-targeted adversarial examples
are easy to find, targeted adversarial examples generated using existing
approaches almost never transfer with their target labels. Therefore, we propose novel ensemble-based approaches to generating
transferable adversarial examples. Using such approaches, we observe a large
proportion of targeted adversarial examples that are able to transfer with their target labels for the first time. We also present some geometric studies to help understanding
the transferable adversarial examples. Finally, we show that the adversarial
examples generated using ensemble-based approaches can successfully attack \clarifai,
which is a black-box image classification system.
\end{abstract}

\section{Introduction}
\label{sec:intro}

Recent research has demonstrated that for a deep architecture, it is easy to
generate adversarial examples, which are close to the original ones but are
misclassified by the deep architecture~(\cite{intriguing, fast-gradient-sign}).
The existence of such adversarial examples may have severe consequences,
which hinders vision-understanding-based applications, such as autonomous driving.
Most of these studies require explicit knowledge of the underlying models.
It remains an open question how to efficiently find adversarial examples
for a black-box model.

Several works have demonstrated that some adversarial examples generated for one model
may also be misclassified by another model. Such a property is referred to
as \emph{transferability}, which can be leveraged to perform black-box attacks. 
This property has been exploited by constructing a substitute of the black-box model,
and generating adversarial instances against the substitute to attack the black-box
system~(\cite{papernot2016transferability,papernot2016practical}).
However, so far, transferability is mostly examined over small datasets, such as
MNIST~(\cite{mnist}) and CIFAR-10~(\cite{cifar}). It has yet to be better understood
transferability over large scale datasets, such as ImageNet~(\cite{imagenet}).

In this work, we are the first to conduct an extensive study of the transferability of different
adversarial instance generation strategies applied to different state-of-the-art models
trained over a large scale dataset. In particular, we study two types of adversarial
examples: (1) non-targeted adversarial examples, which can be misclassified by a
network, regardless of what the misclassified labels may be; and (2)
targeted adversarial examples, which can be classified by a network as a target label.
We examine several existing approaches searching for adversarial examples based on a
single model. While non-targeted adversarial examples are more likely to
transfer, we observe few targeted adversarial examples that are able to transfer with their target labels.

We further propose a novel strategy to generate transferable adversarial images
using an ensemble of multiple models. In our evaluation, we observe that this new
strategy can generate non-targeted adversarial instances with better transferability
than other methods examined in this work. Also, for the first time, we observe a large
proportion of targeted adversarial examples that are able to transfer with their target labels.

We study geometric properties of the models in our evaluation. In particular, we 
show that the gradient directions of different models are orthogonal to each other. 
We also show that decision boundaries of different models align well with each other,
which partially illustrates why adversarial examples can transfer.

Last, we study whether generated adversarial images can attack
\clarifai, a commercial company providing state-of-the-art image classification
services. We have no knowledge about the training dataset and the types of models 
used by \clarifai; meanwhile, the label set of \clarifai is quite different from 
ImageNet's. We show that even in this case, both non-targeted and targeted
adversarial images transfer to \clarifai. This is the first work documenting the success of generating
both non-targeted and targeted adversarial examples for a black-box state-of-the-art online
image classification system, whose model and training dataset are unknown to the attacker.

\paragraph{Contributions and organization.} We summarize our main contributions as follows:
\begin{itemize}
    \item For ImageNet models, we show that while existing approaches are effective to generate non-targeted transferable adversarial examples (Section~\ref{sec:non-targeted}), only few targeted adversarial examples generated by existing methods can transfer (Section~\ref{sec:targeted}).
    \item We propose novel ensemble-based approaches to generate adversarial examples (Section~\ref{sec:ensemble}). Our approaches enable a large portion of targeted adversarial examples to transfer among multiple models for the first time.
    \item We are the first to present that targeted adversarial examples generated for models trained on ImageNet can transfer to a black-box system, i.e., \clarifai, whose model, training data, and label set is unknown to us  (Section~\ref{sec:real}). In particular, \clarifai's label set is very different from ImageNet's.
    \item We conduct the first analysis of geometric properties for large models trained over ImageNet (Section~\ref{sec:geo}), and the results reveal several interesting findings, such as the gradient directions of different models are orthogonal to each other.
\end{itemize}
In the following, we first discuss related work, and then present the background knowledge and
experiment setup in Section~\ref{sec:adl}. Then we present each of our experiments and conclusions in the corresponding section as mentioned above.

\paragraph{Related work.}
Transferability of adversarial examples
was first examined by~\cite{intriguing}, which studied the transferability
(1) between different models trained over the same dataset; and
(2) between the same or different model trained over disjoint subsets of a dataset; 
However, \cite{intriguing} only studied MNIST.

\eat{
\xinyun{I don't think this summary is completely correct. For MNIST, they did experiments using FC (fully-connected network) and AE (classifier on top of an AutoEncoder). By saying different hyperparameters, they mean a model architecture with different number of hidden units: for example, FC100-100-10 and FC123-456-10 have different hyperparameters. However, I don't think they are the same model, since their model architectures are actually different. And they studied transferability between FC and AE as well, which are entirely different model architectures. In addition, they also studied transferability between different models trained over different datasets, but it is not pointed out in the summary.

Another thing is, actually this paper also studied adversarial examples on other datasets such as ImageNet, although it didn't study transferability on ImageNet.

Therefore, I think the following summary may be more appropriate:
(1) between different models trained over the same dataset;
(2) between the same model trained over different datasets, i.e., disjoint subsets of MNIST.
(3) between different models trained over different datasets.

However, \cite{intriguing} only studied transferability on MNIST.
}}

The study of transferability was followed by~\cite{fast-gradient-sign}, which attributed the phenomenon of
transferability to the reason that the adversarial perturbation is highly aligned with
the weight vector of the model. Again, this hypothesis was tested using MNIST and CIFAR-10 datasets. We show that this is not the case for models trained over ImageNet.

\eat{
\cite{papernot2016transferability, papernot2016practical} examined
constructing a substitute model to attack a black-box target model. To train the substitute model, they developed a technique that synthesizes a training set and annotates it by querying the target model for labels. In particular, they demonstrate that black-box attacks are feasible towards machine learning services hosted by Amazon, Google and MetaMind. In these works, however, only the model and the training process are a black box, but the training and test sets are controlled by the attacker. In contrast, we attack \clarifai, whose model, training data, training process, and even the test label set are unknown to the attacker.

Further, ~\cite{papernot2016transferability} studied the transferability between deep neural
networks and other models such as decision tree, kNN, etc. Nevertheless, the datasets studied in these
works are also small scale, i.e., MNIST and GTSRB~(\cite{gtsrb}). In this work, we study
the transferability over larger models and a larger dataset without constructing the substitute model ourselves.
}

\cite{papernot2016transferability, papernot2016practical} examined
constructing a substitute model to attack a black-box target model. To train the substitute model, they developed a technique that synthesizes a training set and annotates it by querying the target model for labels. They demonstrate that using this approach, black-box attacks are feasible towards machine learning services hosted by Amazon, Google, and MetaMind. Further, ~\cite{papernot2016transferability} studied the transferability between deep neural networks and other models such as decision tree, kNN, etc. 

Our work differs from \cite{papernot2016transferability, papernot2016practical} in three aspects. First, in these works, only the model and the training process are a black box, but the training set and the test set are controlled by the attacker; in contrast, we attack \clarifai, whose model, training data, training process, and even the test label set are unknown to the attacker. Second, the datasets studied in these works are small scale, i.e., MNIST and GTSRB~(\cite{gtsrb}); in our work, we study the transferability over larger models and a larger dataset, i.e., ImageNet. Third, to attack black-box machine learning systems, we do not query the systems for constructing the substitute model ourselves.

In a concurrent and independent work, \cite{moosavi2016universal} showed the 
existence of a \emph{universal perturbation} for each model, which can transfer 
across different images. They also show that the adversarial images generated 
using these universal perturbations can transfer across different models on ImageNet.
However, they only examine the non-targeted transferability, while our work studies
both non-targeted and targeted transferability over ImageNet.


\section{Adversarial Deep Learning and Transferability}
\label{sec:adl}

\subsection{The adversarial deep learning problem}
\label{sec:back}

We assume a classifier $f_\theta(x)$ outputs a category (or a label)
as the prediction. Given an original image $x$, with ground truth label
$y$, the adversarial deep learning problem is to seek for 
\emph{adversarial examples} for the classifier $f_\theta(x)$. Specifically,
we consider two classes of adversarial examples. A \emph{non-targeted} adversarial 
example $x^\star$ is an instance that is close to $x$, in which case $x^\star$ 
should have the same ground truth as $x$, while $f_\theta(x^\star)\neq y$. For
the problem to be non-trivial, we assume $f_\theta(x)=y$ without loss of generality.
A \emph{targeted} adversarial example $x^\star$ is close to $x$ and satisfies 
$f_\theta(x^\star)=y^\star$, where $y^\star$ is a target label specified by the adversary, and $y^\star \neq y$.

\subsection{Approaches for generating adversarial examples}

In this work, we consider three classes of approaches for generating
adversarial examples: optimization-based approaches, fast gradient
approaches, and fast gradient sign approaches. Each class has non-targeted and targeted versions respectively.

\subsubsection{Approaches for generating non-targeted adversarial examples}
Formally, given an image $x$ with ground truth $y=f_\theta(x)$, 
searching for a non-targeted adversarial example can be modeled
as searching for an instance $x^\star$ to satisfy the following constraints:
\begin{eqnarray}
f_\theta(x^\star)\neq y\label{eq:misclass-nt}\\
d(x, x^\star)\leq B\label{eq:dist-nt}
\end{eqnarray}
where $d(\cdot, \cdot)$ is a metric to quantify the distance between an
original image and its adversarial counterpart, and $B$, called \emph{distortion},
is an upper bound placed on this distance. 
Without loss of generality, we consider model
$f$ is composed of a network $J_\theta(x)$, which outputs the probability
for each category, so that $f$ outputs the category with the highest
probability.
%
\paragraph{Optimization-based approach.}
One approach is to approximate the solution to the following
optimization problem:
\begin{equation}
\mathbf{argmin}_{x^\star} \lambda d(x, x^\star) - \ell(\mathbf{1}_y, J_\theta(x^\star))
\label{obj:non-targeted}
\end{equation}
where $\mathbf{1}_y$ is the one-hot encoding of the ground truth label $y$,
$\ell$ is a loss function to measure the distance between the prediction
and the ground truth, and $\lambda$ is a constant to balance
constraints~(\ref{eq:dist-nt}) and~(\ref{eq:misclass-nt}), which is
empirically determined. Here, loss function $\ell$ is used to
approximate constraint~(\ref{eq:misclass-nt}), and its choice can affect
the effectiveness of searching for an adversarial example. In this work,
we choose $\ell(u,v) = \log{(1-u\cdot v)}$, which is shown to be
effective by~\cite{carlini2016towards}.

\paragraph{Fast gradient sign (FGS).} \cite{fast-gradient-sign} proposed
the fast gradient sign (FGS) method so that the gradient needs be computed only
once to generate an adversarial example. FGS can be used to generate
adversarial images to meet the $L_\infty$ norm bound. 
Formally, non-targeted adversarial examples are constructed as
\[x^\star \leftarrow \mathrm{clip}(x + B \mathbf{sgn}(\nabla_x \ell(\mathbf{1}_y, J_\theta(x))))\]
Here, $\mathrm{clip}(x)$ is used to clip each dimension of $x$ to the range
of pixel values, i.e., $[0,255]$ in this work. 
We make a slight variation to choose
$\ell(u,v) = \log{(1-u\cdot v)}$, which is the same as used in the
optimization-based approach.

\paragraph{Fast gradient (FG).} The fast gradient approach (FG) is similar to
FGS, but instead of moving along the gradient sign direction, FG moves along
the gradient direction. In particular, we have 
\[x^\star \leftarrow \mathrm{clip}(x + B \frac{\nabla_x \ell(\mathbf{1}_y, J_\theta(x))}{||\nabla_x \ell(\mathbf{1}_y, J_\theta(x))||}))\]
Here, we assume the distance metric in constraint (\ref{eq:dist-nt}),
$d(x, x^\star)=||x - x^\star||$ is a norm of $x-x^\star$.
The term $\mathbf{sgn}(\nabla_x \ell)$ in FGS is replaced by
$\frac{\nabla_x \ell}{||\nabla_x \ell||}$ to meet this distance constraint.

We call both FGS and FG \emph{fast gradient-based approaches}.

\subsubsection{Approaches for generating targeted adversarial examples}

A targeted adversarial image $x^\star$ is similar to a non-targeted one, but constraint~(\ref{eq:misclass-nt}) is replaced by
\begin{eqnarray}
f_\theta(x^\star)=y^\star\label{eq:misclass-t}
\end{eqnarray}
where $y^\star$ is the target label given by the adversary.
For the optimization-based approach, we approximate the solution by solving
the following dual objective:
\begin{equation}
\mathbf{argmin}_{x^\star} \lambda d(x, x^\star) + \ell'(\mathbf{1}_{y^\star}, J_\theta(x^\star))
\label{obj:targeted}
\end{equation}
In this work, we choose the standard cross entropy loss $\ell'(u,v) = -\sum \limits_i u_i \log v_i$.

For FGS and FG, we construct adversarial examples as follows:
\[x^\star \leftarrow \mathrm{clip}(x - B \mathbf{sgn}(\nabla_x \ell'(\mathbf{1}_{y^\star}, J_\theta(x))))~~~~\mathrm{(FGS)}\]
\[x^\star \leftarrow \mathrm{clip}(x - B \frac{\nabla_x \ell'(\mathbf{1}_{y^\star}, J_\theta(x))}{||\nabla_x \ell'(\mathbf{1}_{y^\star}, J_\theta(x))||})~~~~\mathrm{(FG)}\]
where $\ell'$ is the same as the one used for the optimization-based approach.

\subsection{Evaluation Methodology}
\label{sec:setup}

For the rest of the paper, we focus on examining the transferability
among state-of-the-art models trained over ImageNet~(\cite{imagenet}).
In this section, we detail the models to be examined, the dataset to be evaluated,
and the measurements to be used.

\paragraph{Models.} We examine five networks,
ResNet-50, ResNet-101, ResNet-152~(\cite{he2015deep})\footnote{\scriptsize{\url{https://github.com/KaimingHe/deep-residual-networks}}},
GoogLeNet~(\cite{googlenet})\footnote{\scriptsize{\url{https://github.com/BVLC/caffe/tree/master/models/bvlc_googlenet}}}, and
VGG-16~(\cite{vgg16})\footnote{\scriptsize{\url{https://gist.github.com/ksimonyan/211839e770f7b538e2d8}}}.
We retrieve the pre-trained models for each network online. The performance of these models on the ILSVRC 2012~(\cite{imagenet}) validation set 
\ifarxiv
are presented in the appendix (Table~\ref{tab:accuracy}). 
\else
can be found in our online technical report~\chang{cite tr}.
\fi
We choose these models to study the transferability
between homogeneous architectures (i.e., ResNet models) and heterogeneous architectures.

\paragraph{Dataset.} It is less meaningful to examine the transferability of
an adversarial image between two models which cannot classify the original
image correctly. Therefore, from the ILSVRC 2012 validation set, we randomly
choose 100 images, which can be classified correctly by all five models in our examination. \eat{We also manually inspect the labels to enforce that (1) no two images
share the same ground truth label; and (2) the semantics of the ground truth
labels are as diverse as possible. }These 100 images form
our test set. To perform targeted attacks, we manually choose a target label for each image, so that
its semantics is far from the ground truth. The
images and target labels in our evaluation can be found on website\footnote{\scriptsize{\url{https://github.com/sunblaze-ucb/transferability-advdnn-pub}}}.

\paragraph{Measuring transferability.} Given two models,
we measure the non-targeted transferability by computing the percentage
of the adversarial examples generated for one model that can be classified
correctly for the other. We refer to this percentage as \emph{accuracy}.
A lower accuracy means better non-targeted transferability.
We measure the targeted transferability by computing the percentage of
the adversarial examples generated for one model that are classified as
the target label by the other model.
We refer to this percentage as \emph{matching rate}.
A higher matching rate means better targeted transferability.
For clarity, the reported results
are only based on top-1 accuracy. Top-5 accuracy's counterparts can be found in the appendix.

\paragraph{Distortion.} Besides transferability, another important factor
is the distortion between adversarial images and the original ones.
We measure the distortion by
\emph{root mean square deviation}, i.e., RMSD, which is computed as
$d(x^\star, x) = \sqrt{\sum_i(x_i^\star - x_i)^2/N}$,
where $x^\star$ and $x$ are the vector representations of an adversarial image
and the original one respectively, $N$ is the dimensionality of $x$ and $x^\star$,
and $x_i$ denotes the pixel value of the $i$-th dimension of $x$, 
within range $[0, 255]$, and similar for $x^\star_i$.

\section{Non-targeted Adversarial Examples}
\label{sec:non-targeted}

In this section, we examine different approaches for generating
non-targeted adversarial images.

\subsection{Optimization-based approach}

To apply the optimization-based approach for a single model,
we initialize $x^\star$ to be $x$ and use Adam Optimizer~(\cite{adam})
to optimize Objective~(\ref{obj:non-targeted})
. We find that we can tune the RMSD by adjusting the learning rate of
Adam and $\lambda$. We find that, for each model,
we can use a small learning rate
to generate adversarial images with small RMSD, i.e. $<2$,
with any $\lambda$. In fact, we find that when
initializing $x^\star$ with $x$, Adam Optimizer will search for
an adversarial example around $x$, even when we set $\lambda$
to be $0$, i.e., not restricting the distance between $x^\star$ and
$x$. Therefore, we set $\lambda$ to be $0$ for all experiments using optimization-based approaches throughout the paper.
Although these adversarial examples with small distortions can successfully
fool the target model, however, they cannot transfer well to other models
(see Table~\ref{tab:top1-nontargeted-o-small} and~\ref{tab:top5-nontargeted-o-small} in the appendix for details).

We increase the learning rate to allow
the optimization algorithm to search for adversarial images with larger
distortion. In particular, we set the learning rate to be $4$.
We run Adam Optimizer for 100 iterations to generate the adversarial images.
We observe that the loss converges after 100 iterations.
An alternative optimization-based approach leading to similar results can
be found in the appendix.

\paragraph{Non-targeted adversarial examples transfer.}
We generate
non-targeted adversarial examples on one network, but evaluate them on
another, and Table~\ref{tab:non-targeted-opt} Panel A presents the results.
From the table, we can observe that
\begin{itemize}
\item The diagonal contains all 0 values. This says that all
    adversarial images generated for one model can mislead the same model.
\item A large proportion of non-targeted adversarial images generated for one
    model using the optimization-based approach can transfer to another.
\item Although the three ResNet models share similar 
    architectures which differ only in the hyperparameters, adversarial examples generated against a ResNet model do not necessarily transfer to another ResNet model better than other non-ResNet models. For example,
    the adversarial examples generated for VGG-16 have lower accuracy on ResNet-50
    than those generated for ResNet-152 or ResNet-101.
\end{itemize}

\eat{
Notice that an alternative method to generate non-targeted adversarial
examples with large distortion is to revise the optimization objective to
incorporate this distortion constraint.
For example, for non-targeted adversarial image searching, we can optimize
for the following objective.
\[\mathbf{argmin}_{x^\star} -\log{(1-\mathbf{1}_y \cdot J_\theta(x^\star))} + \lambda_1 \mathrm{ReLU}(\tau - d(x, x^\star)) + \lambda_2 \mathrm{ReLU}(d(x, x^\star) - \tau)\]
Optimizing for the above objective has the following three effects:
(1) minimizing $-\log{(1-\mathbf{1}_y \cdot J_\theta(x^\star))}$;
(2) Penalizing the solution if $d(x, x^\star)$ is no more than a threshold $\tau$ (too low); and
(3) Penalizing the solution if $d(x, x^\star)$ is too high.

In our preliminary evaluation, we found that the solutions computed from
the two approaches have similar transferability. We thus omit the results
for this alternative approach.
}

\begin{table}[t]
\centering
\begin{tabular}{|c|c|c|c|c|c|c|c|}
\hline
                  & RMSD  & \small ResNet-152 & \small ResNet-101 & \small ResNet-50 & \small VGG-16 & \small GoogLeNet \\ \hline
\small ResNet-152 & 22.83 & 0\%               & 13\%              & 18\%             & 19\%          & 11\%             \\ \hline
\small ResNet-101 & 23.81 & 19\%              & 0\%               & 21\%             & 21\%          & 12\%             \\ \hline
\small ResNet-50  & 22.86 & 23\%              & 20\%              & 0\%              & 21\%          & 18\%             \\ \hline
\small VGG-16     & 22.51 & 22\%              & 17\%              & 17\%             & 0\%           & 5\%              \\ \hline
\small GoogLeNet  & 22.58 & 39\%              & 38\%              & 34\%             & 19\%          & 0\%                           \\ \hline
\multicolumn{7}{c}{Panel A: Optimization-based approach}
\\
\multicolumn{7}{c}{}
\\
\hline
                  & \small RMSD  & \small ResNet-152 & \small ResNet-101 & \small ResNet-50 & \small VGG-16 & \small GoogLeNet 
                  \\ \hline
\small ResNet-152 & 23.45 & 4\%               & 13\%              & 13\%             & 20\%          & 12\%           
\\ \hline
\small ResNet-101 & 23.49 & 19\%              & 4\%               & 11\%             & 23\%          & 13\%            
\\ \hline
\small ResNet-50  & 23.49 & 25\%              & 19\%              & 5\%              & 25\%          & 14\%             
\\ \hline
\small VGG-16     & 23.73 & 20\%              & 16\%              & 15\%             & 1\%           & 7\%              
\\ \hline
\small GoogLeNet  & 23.45 & 25\%              & 25\%              & 17\%             & 19\%          & 1\%              
\\ \hline
\multicolumn{7}{c}{Panel B: Fast gradient approach}
\end{tabular}
\caption{Transferability of non-targeted adversarial images generated between pairs of models. 
The first column indicates the average RMSD of all
adversarial images generated for the model in the corresponding row. The cell $(i, j)$ indicates the accuracy of the
adversarial images generated for model $i$ (row) evaluated over model $j$ (column). Results of top-5 accuracy can be found in the appendix (Table~\ref{tab:top5-non-targeted-opt} and Table~\ref{tab:top5-trans-fg}).}
\label{tab:non-targeted-opt}
\end{table}

\subsection{Fast gradient-based approaches}

We then examine the effectiveness of fast gradient-based approaches.
A good property of fast gradient-based approaches is that all generated
adversarial examples lie in a 1-D subspace. Therefore, we can easily
approximate the minimal distortion in this subspace of transferable
adversarial examples between two models. In the following, we first 
control the RMSD to study fast gradient-based approaches' effectiveness.
Second, we study the transferable minimal distortions of fast gradient-based 
approaches.

\subsubsection{Effectiveness and transferability of the fast gradient-based approaches}

Since the distortion $B$ and the RMSD of the generated adversarial images
are highly correlated, we can choose this hyperparameter $B$ to generate
adversarial images with a given RMSD. In Table~\ref{tab:non-targeted-opt} Panel B, 
we generate adversarial images using FG such that the average RMSD is almost
the same as those generated using the optimization-based approach.
We observe that the diagonal values in the table are
all positive, which means that FG cannot fully mislead the models.
A potential reason is that, FG can be viewed as approximating the optimization, but is tailored for speed over accuracy.

On the other hand, the values of non-diagonal cells in
the table, which correspond to the accuracies of adversarial
images generated for one model but evaluated on another, are comparable with or
less than their counterparts in the optimization-based approach.
This shows that non-targeted adversarial examples generated by FG
exhibit transferability as well.

We also evaluate FGS, but the transferability of the generated
images is worse than the ones generated using either FG or
optimization-based approaches. The
results are in the appendix (Table~\ref{tab:trans-fgs} and~\ref{top5-trans-fgs}). It shows that
when RMSD is around 23, the accuracies of the adversarial images
generated by FGS is greater than their counterparts for FG.
We hypothesize the reason why transferability of FGS is worse
to this fact.


\subsubsection{Adversarial images with minimal transferable RMSD}

For an image $x$ and two models $M_1, M_2$, we can approximate the minimal
distortion $B$ along a direction $\delta$, such that $x_B=x+B\delta$ generated
for $M_1$ is adversarial for both $M_1$ and $M_2$. Here $\delta$ is the
direction, i.e., $\mathbf{sgn}(\nabla_x\ell)$ for FGS,
and $\nabla_x\ell/||\nabla_x\ell||$ for FG.

We refer to \emph{the minimal transferable RMSD from $M_1$ to $M_2$
	using FG (or FGS)} as the RMSD of a
transferable adversarial example $x_B$ with the minimal transferable
distortion $B$ from $M_1$ to $M_2$ using FG (or FGS).
The minimal transferable RMSD can illustrate the tradeoff
between distortion and transferability.

In the following, 
we approximate the minimal transferable RMSD through
a linear search by sampling $B$ every 0.1 step.
We choose the linear-search method rather than 
binary-search method to determine the minimal
transferable RMSD because the adversarial images generated
from an original image may come from multiple intervals.
The experiment can be found in the appendix (Figure~\ref{fig:non-contiguous}).

\paragraph{Minimal transferable RMSD using FG and FGS.}

\begin{figure}[t]
\begin{subfigure}{0.5\linewidth}
\centering
  \includegraphics[scale=0.25]{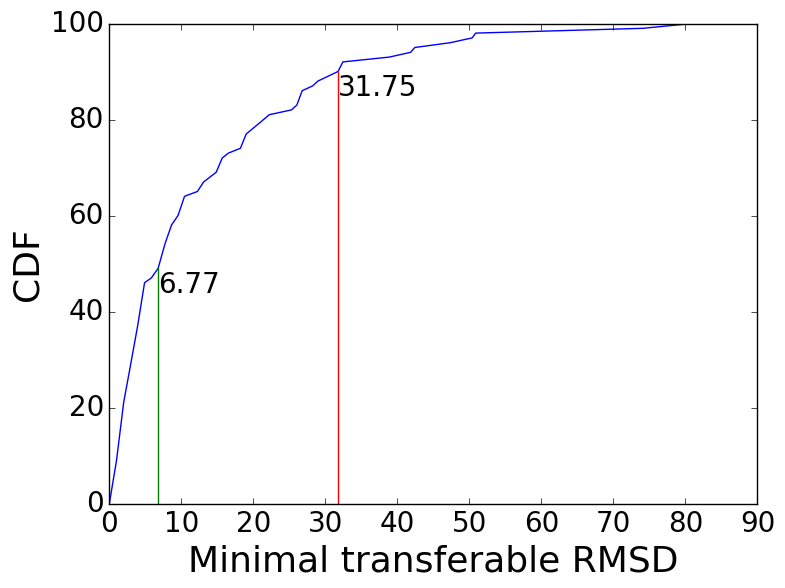}
  \caption{Fast Gradient}
  \label{fig:tradeoff-fg}
\end{subfigure}
\begin{subfigure}{0.5\linewidth}
\centering
  \includegraphics[scale=0.25]{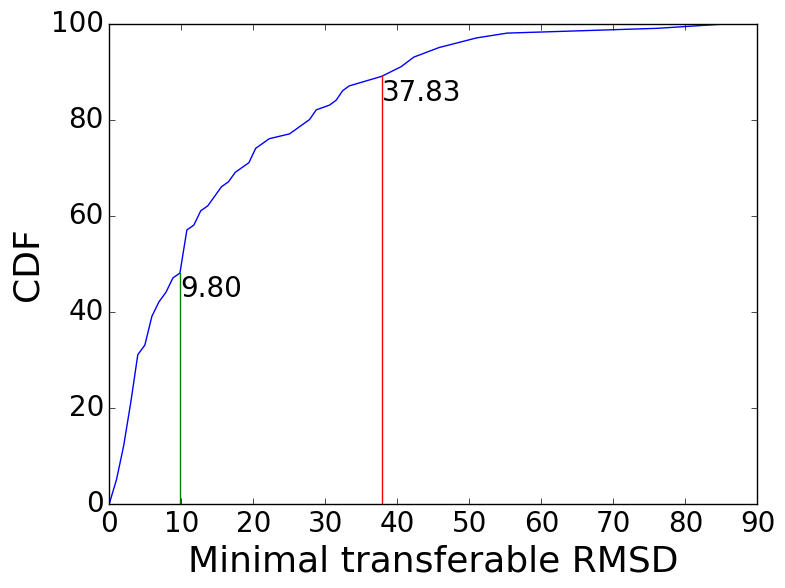}
  \caption{Fast Gradient Sign}
  \label{fig:tradeoff-fgs}
\end{subfigure}
  \caption{The CDF of the minimal transferable RMSD
  from VGG-16 to ResNet-152 using FG (a) and FGS (b). The green line labels the median minimal
  transferable RMSD, while the red line labels the minimal transferable
  RMSD to reach 90\% percentage.}
  \label{fig:tradeoff}
\end{figure}

Figure~\ref{fig:tradeoff} plots the cumulative distribution
function (CDF) of the minimal
transferable RMSD from VGG-16 to ResNet-152 using non-targeted
FG (Figure~\ref{fig:tradeoff-fg}) and FGS (Figure~\ref{fig:tradeoff-fgs}).
From the figures, we observe that both FG and FGS can find
100\% transferable adversarial images with RMSD less than $80.91$
and $86.56$ respectively.
Further, the FG method can generate transferable attacks with
smaller RMSD than FGS. A potential reason is that while FGS
minimizes the distortion's
$L_\infty$ norm, FG minimizes its $L_2$ norm, which is proportional to RMSD.

\subsection{Comparison with random perturbations}
\label{sec:non-target-noise}

We also evaluate the test accuracy when we add a Gaussian noise to the 100 images in our test set. The concrete results can be found in the appendix, and we show the conclusion that the ``transferability" of this approach is significantly worse than either optimization-based approaches or fast gradient-based approaches.

\section{Targeted Adversarial Examples}
\label{sec:targeted}

In this section, we examine the transferability of targeted adversarial images.
Table~\ref{tab:matching-target-opt} presents the results for using optimization-based
approach.
We observe that (1) the prediction of targeted adversarial images can match
the target labels when evaluated on the same model that is used to generate the adversarial examples;
but (2) the targeted adversarial images can be rarely predicted as the target labels
by a different model. We call the latter that \emph{the target labels do not transfer}.
Even when we increase the distortion, we still do not observe
improvements on making target label transfer. Some results can be found in the appendix
(Table~\ref{tab:matching-target-opt-large}). Even if we compute the matching rate based on
top-5 accuracy, the highest matching rate is only 10\%. The results can be found in the
appendix (Table~\ref{tab:top5-matchingrate}).

\begin{table}[t]
\centering
\begin{tabular}{|c|c|c|c|c|c|c|c|}
\hline
           & \small RMSD & \small ResNet-152 & \small ResNet-101 & \small ResNet-50 & \small VGG-16 & \small GoogLeNet 
           \\ \hline
\small ResNet-152 & 23.13 & 100\%      & 2\%        & 1\%       & 1\%    & 1\%      
\\ \hline
\small ResNet-101 & 23.16 & 3\%        & 100\%      & 3\%       & 2\%    & 1\%    
 \\ \hline
\small ResNet-50  & 23.06 & 4\%        & 2\%        & 100\%     & 1\%    & 1\%    
     \\ \hline
\small VGG-16     & 23.59 & 2\%        & 1\%        & 2\%       & 100\%  & 1\%     
      \\ \hline
\small GoogLeNet  & 22.87 & 1\%        & 1\%        & 0\%       & 1\%    & 100\%     
\\ \hline
\end{tabular}
\caption{The matching rate of targeted adversarial images generated using the optimization-based approach.
The first column indicates the average RMSD of the generated adversarial images.
Cell $(i, j)$ indicates that matching rate of the targeted adversarial images
generated for model $i$ (row) when evaluated on model $j$ (column). The top-5 results can be found in the appendix (Table~\ref{tab:top5-matchingrate-opt}).}
\label{tab:matching-target-opt}
\end{table}

We also examine the targeted adversarial images generated by
fast gradient-based approaches, and we observe that the target labels do not transfer as well. The results are
deferred to the appendix (Table~\ref{tab:fg-targeted}).
In fact, most targeted adversarial images cannot
mislead the model, for which the adversarial images are generated,
to predict the target labels, regardless of how large the distortion is used.
We attribute it to the fact that the fast gradient-based
approaches only search for attacks in a 1-D subspace. In this subspace,
the total possible predictions
may contain a small subset of all labels, which usually does not contain
the target label.
In Section~\ref{sec:geo}, we study decision boundaries regarding this issue.
\changcomment{We may do exp to validate it: Therefore, for most of the targeted
labels, it is impossible for the generated attacks meet the targeted labels.}

\ifshort
\else
We also evaluate the matching rate of images added with Gaussian noise, as
described in Section~\ref{sec:non-target-noise}. However, we observe that the matching rate of any of the 5 models is $0\%$.
Therefore, we conclude that by adding Gaussian noise, the attacker cannot generate successful targeted adversarial examples at all, 
let alone targeted transferability.
\fi
\section{Ensemble-based approaches}
\label{sec:ensemble}

We hypothesize that if an adversarial image remains adversarial
for multiple models, then it is more likely to transfer
to other models as well. We develop techniques to generate
adversarial images for multiple models. The basic idea is to
generate adversarial images for \emph{the ensemble of the models}.
Formally, given $k$ white-box models with softmax outputs being 
$J_1,...,J_k$, an original image $x$,
and its ground truth $y$, \emph{the ensemble-based approach} solves the
following optimization problem (for targeted attack):
\begin{equation}
\mathbf{argmin}_{x\star} -\log{\big((\sum_{i=1}^k \alpha_iJ_i(x^\star))\cdot\mathbf{1}_{y^\star}\big)} + \lambda d(x, x^\star)\label{eq:opt-ensemble}
\end{equation}
where $y^\star$ is the target label specified by the adversary,
$\sum\alpha_iJ_i(x^\star)$ is the
ensemble model, and $\alpha_i$ are the ensemble weights,
$\sum_{i=1}^{k}\alpha_i = 1$.
Note that (\ref{eq:opt-ensemble}) is the targeted objective.
The non-targeted counterpart can be derived similarly.
In doing so, we hope the generated adversarial images remain
adversarial for an additional black-box model $J_{k+1}$.

We evaluate the effectiveness of the ensemble-based approach.
For each of the five models, we treat it as the black-box model
to attack, and generate adversarial images
for the ensemble of the rest four, which is considered as white-box.
We evaluate the generated adversarial images over all five models.
Throughout the rest of the paper, we refer to the approaches evaluated
in Section~\ref{sec:non-targeted} and~\ref{sec:targeted} as the
approaches using a single model, and to the ensemble-based approaches
discussed in this section as the approaches using an ensemble model.

\paragraph{Optimization-based approach.}
We use Adam to optimize the objective (\ref{eq:opt-ensemble}) with
equal ensemble weights across all models in the ensemble to
generate targeted adversarial examples. In particular, we set the learning
rate of Adam to be $8$ for each model. In each iteration,
we compute the Adam update for each model, sum up the four updates,
and add the aggregation onto the image. We
run 100 iterations of updates, and we observe that the loss
converges after 100 iterations.
By doing so, for the first time, we observe a large proportion
of the targeted adversarial images whose target labels can transfer.
The results are presented in Table~\ref{tab:t-trans-label}.
We observe that not all targeted adversarial images can
be misclassified to the target labels by the models used in
the ensemble. This suggests that while
searching for an adversarial example for the ensemble model,
there is no direct supervision to mislead any individual model in the ensemble
to predict the target label. Further, from the diagonal numbers of the table, we observe that the transferability to ResNet models
is better than to VGG-16 or GoogLeNet, when adversarial examples are generated against
all models except the target model. 

\begin{table}[t]
\centering
\begin{tabular}{|c|c|c|c|c|c|c|}
\hline
                 & RMSD  & \small ResNet-152 & \small ResNet-101 & \small ResNet-50 & \small VGG-16 & \small GoogLeNet \\ \hline
\small -ResNet-152 & 30.68 & 38\%             & 76\%             & 70\%            & 97\%         & 76\%             \\ \hline
\small -ResNet-101 & 30.76 & 75\%             & 43\%             & 69\%            & 98\%         & 73\%             \\ \hline
\small -ResNet-50  & 30.26 & 84\%             & 81\%             & 46\%            & 99\%         & 77\%             \\ \hline
\small -VGG-16     & 31.13 & 74\%             & 78\%             & 68\%            & 24\%         & 63\%             \\ \hline
\small -GoogLeNet & 29.70 & 90\%             & 87\%             & 83\%            & 99\%         & 11\%             \\ \hline
\end{tabular}
\caption{The matching rate of targeted adversarial images generated using the optimization-based approach. The first column indicates the average RMSD of the generated adversarial images.
Cell $(i, j)$ indicates that percentage of the targeted adversarial images generated
for the ensemble of the four models except model $i$ (row) is predicted as the target
label by model $j$ (column). In each row, the minus sign ``$-$" indicates that the model of
the row is not used when generating the attacks. Results of top-5 matching rate can be found in the appendix (Table~\ref{tab:top5-matchingrate-ensemble}).}
\label{tab:t-trans-label}
\end{table}

We also evaluate non-targeted adversarial images generated by the ensemble-based approach.
We observe that the generated adversarial images
have almost perfect transferability. We use the same procedure as for
the targeted version, except the objective
to generate the adversarial images. We evaluate the generated
adversarial images over all models.
The results are presented in Table~\ref{tab:non-targeted-ensemble-32}.
The generated adversarial images all have RMSDs around 17,
which are lower than 22 to 23 of the optimization-based approach using a single model (See Table~\ref{tab:non-targeted-opt} for comparison).
When the adversarial images are evaluated over models which are not
used to generate the attack, the accuracy is no greater than $6\%$.
For a reference,
the corresponding accuracies for all approaches evaluated in Section~\ref{sec:non-targeted} using one single
model are at least 12\%. Our experiments demonstrate that the ensemble-based
approaches can generate almost perfectly transferable adversarial
images.

\begin{table}[t]
\centering
\begin{tabular}{|c|c|c|c|c|c|c|}
\hline
                 & RMSD  & \small ResNet-152 & \small ResNet-101 & \small ResNet-50 & \small VGG-16 & \small GoogLeNet \\ \hline
\small -ResNet-152 & 17.17 & 0\%              & 0\%              & 0\%             & 0\%          & 0\%              \\ \hline
\small -ResNet-101 & 17.25 & 0\%              & 1\%              & 0\%             & 0\%          & 0\%              \\ \hline
\small -ResNet-50  & 17.25 & 0\%              & 0\%              & 2\%             & 0\%          & 0\%              \\ \hline
\small -VGG-16     & 17.80 & 0\%              & 0\%              & 0\%             & 6\%          & 0\%              \\ \hline
\small -GoogLeNet & 17.41 & 0\%              & 0\%              & 0\%             & 0\%          & 5\%              \\ \hline
\end{tabular}
\caption{Accuracy of non-targeted adversarial images generated using the optimization-based approach. The first column indicates the average RMSD of the generated adversarial images. Cell $(i, j)$ corresponds to the accuracy of the attack generated using
four models except model $i$ (row) when evaluated over model $j$ (column).
In each row, the minus sign ``$-$" indicates that the model of
the row is not used when generating the attacks. Results of top-5 accuracy can be found in the appendix (Table~\ref{tab:top5-non-targeted-ensemble-32}).
}
\label{tab:non-targeted-ensemble-32}
\end{table}

\paragraph{Fast gradient-based approach.} The results for non-targeted fast 
gradient-based approaches applied to the ensemble can be found in the appendix 
(Table~\ref{tab:ensemble-fg-top1},~\ref{tab:ensemble-fg-top5},~\ref{tab:ensemble-fgs-top1} and~\ref{tab:ensemble-fgs-top5}). We observe 
that the diagonal values are not zero, which is the same as we observed in the 
results for FG and FGS applied to a single model. We hypothesize a potential 
reason is that the gradient directions of different models in the ensemble are 
orthogonal to each other, as we will illustrate in Section~\ref{sec:geo}. In 
this case, the gradient direction of the ensemble is almost orthogonal to the 
one of each model in the ensemble. Therefore searching along this direction may 
require large distortion to reach adversarial examples.

For targeted adversarial examples generated using FG and FGS based on an ensemble model, their transferability is
no better than the ones generated using a single model. The results can be found in the appendix (Table~\ref{tab:top1-matchingrate-ensemble-fg},~\ref{tab:top5-matchingrate-ensemble-fg},~\ref{tab:top1-matchingrate-ensemble-fgs} and~\ref{tab:top5-matchingrate-ensemble-fgs}).
We hypothesize the same reason to explain this: there are only few possible target labels in total in the 1-D subspace.


\newcolumntype{R}[2]{%
  >{\adjustbox{angle=#1,lap=\width-(#2)}\bgroup}%
  l%
  <{\egroup}%
}
\newcommand*\rot{\multicolumn{1}{R{90}{1em}}}

\section{Geometric Properties of Different Models}
\label{sec:geo}

In this section, we show some geometric properties of the
models to try to better understand transferable adversarial
examples. Prior works also try to understand the geometic properties
of adversarial examples theoretically~(\cite{fawzi2016robustness}) or
empirically~(\cite{fast-gradient-sign}). In this work, we examine
large models trained over a large dataset with 1000 labels, whose
geometric properties are never examined before. This allows us to make
new observations to better understand the models and their adversarial
examples.

\paragraph{The gradient directions of different models in our evaluation are almost orthogonal to each other.}
%
We study whether the adversarial directions of different
models align with each other. We calculate cosine value of the angle between gradient directions of different models, and the results can be found in the appendix (Table~\ref{tab:cosine-between-models}). 
We observe that all non-diagonal values
are close to 0, which indicates that for most images, their gradient directions with respect to
different models are orthogonal to each other.

\begin{figure}[t]
\centering
  \includegraphics[scale=0.4]{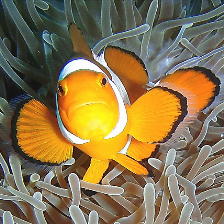}\\
  \caption{The example image to study the decision boundary. Its ID in ILSVRC 2012 validation set
  is 49443, and its ground truth label is ``anemone fish."}
  \label{fig:exmp}
\end{figure}

\begin{figure}[t]
\centering
\begin{tabular}{l@{\hskip -0.06in}c@{\hskip -0.06in}c@{\hskip -0.06in}c@{\hskip -0.06in}c@{\hskip -0.06in}c}
& VGG-16 & ResNet-50 & ResNet-101 & ResNet-152 & GoogLeNet\\
\rot{~~~~~~~~~Zoom-in} &
  \includegraphics[scale=0.28]{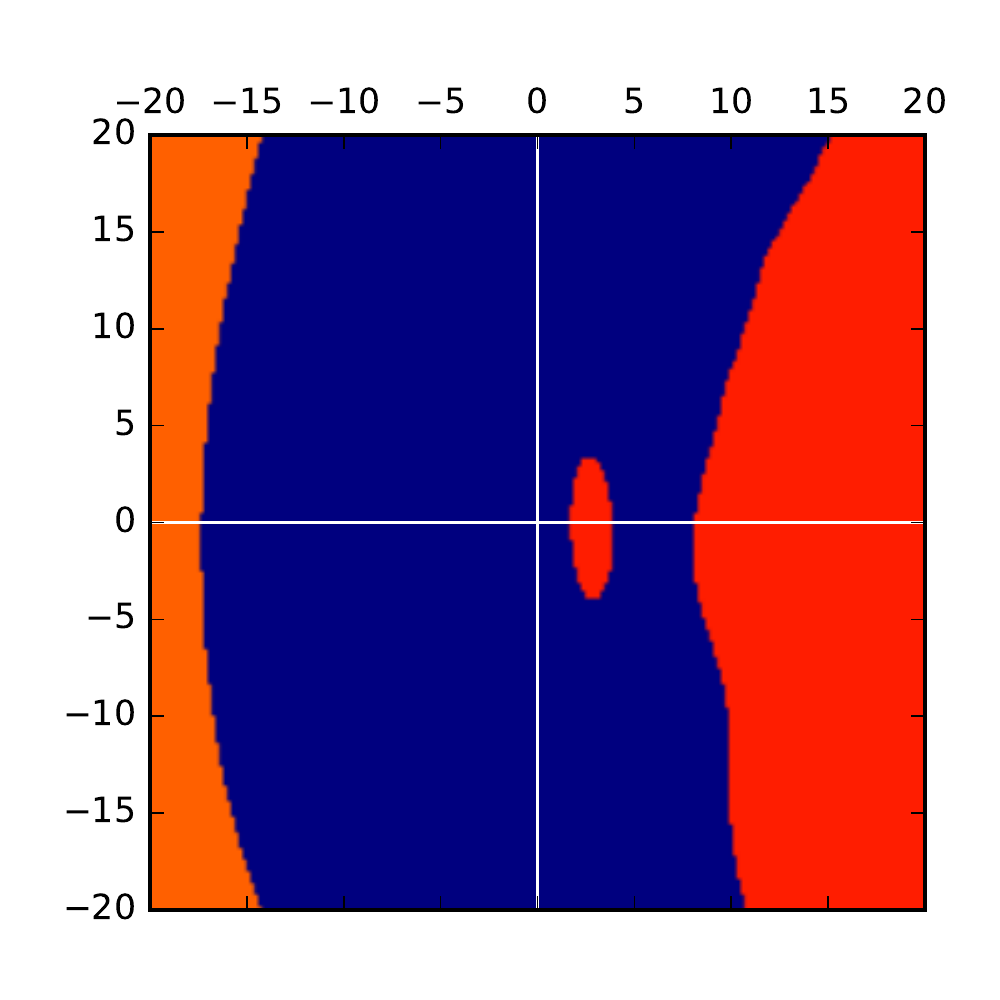} &
  \includegraphics[scale=0.28]{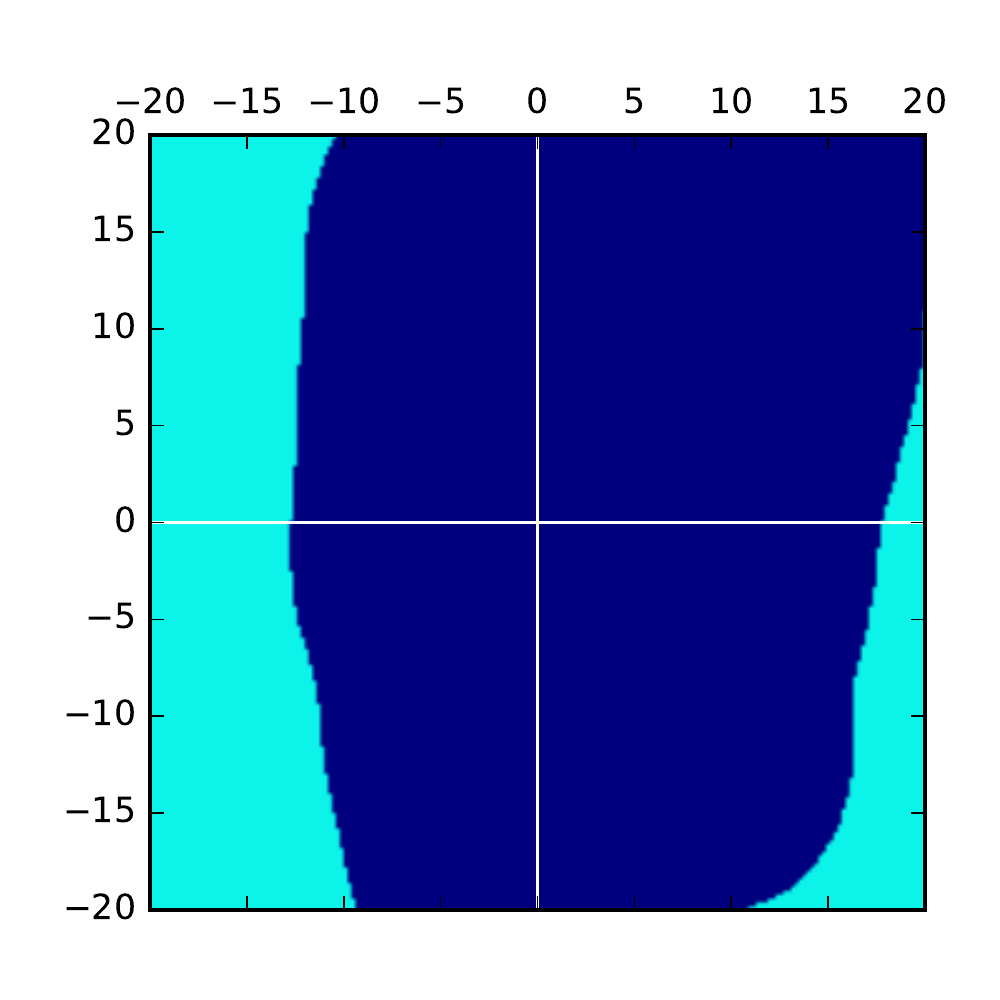} &
  \includegraphics[scale=0.28]{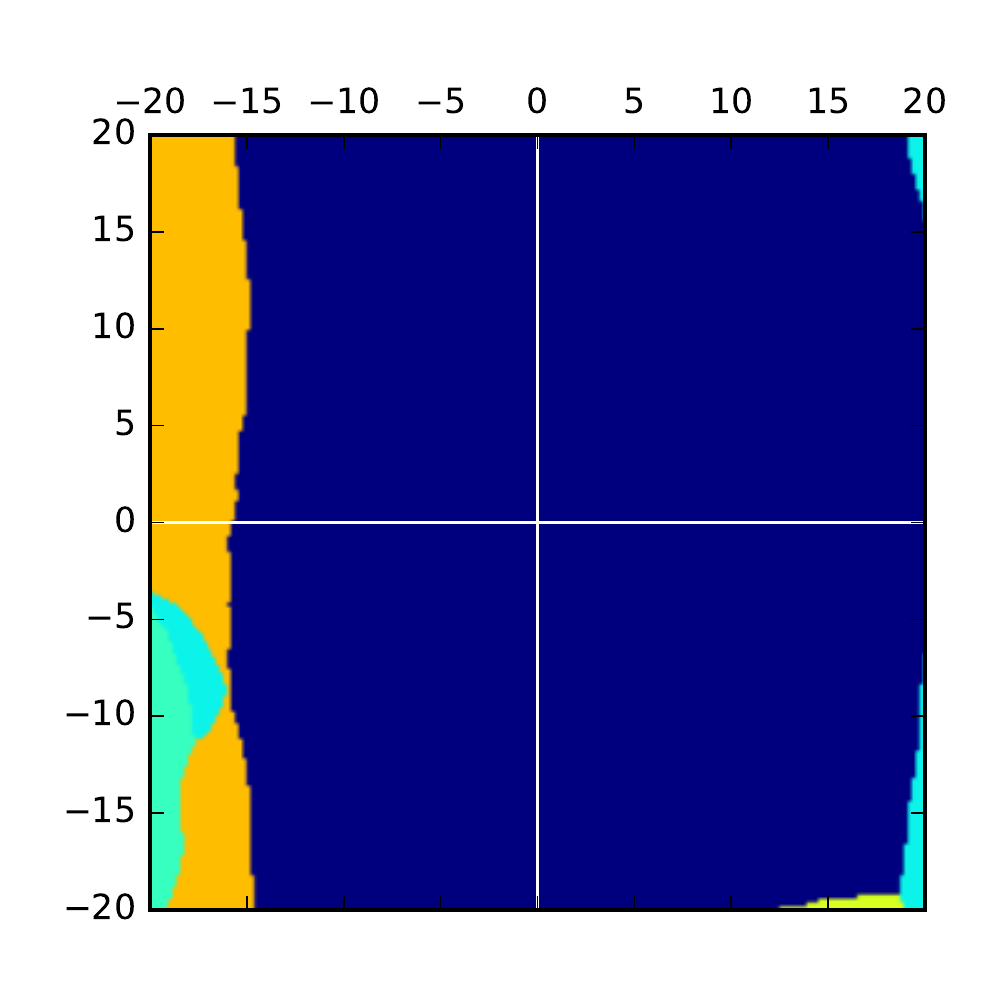} &
  \includegraphics[scale=0.28]{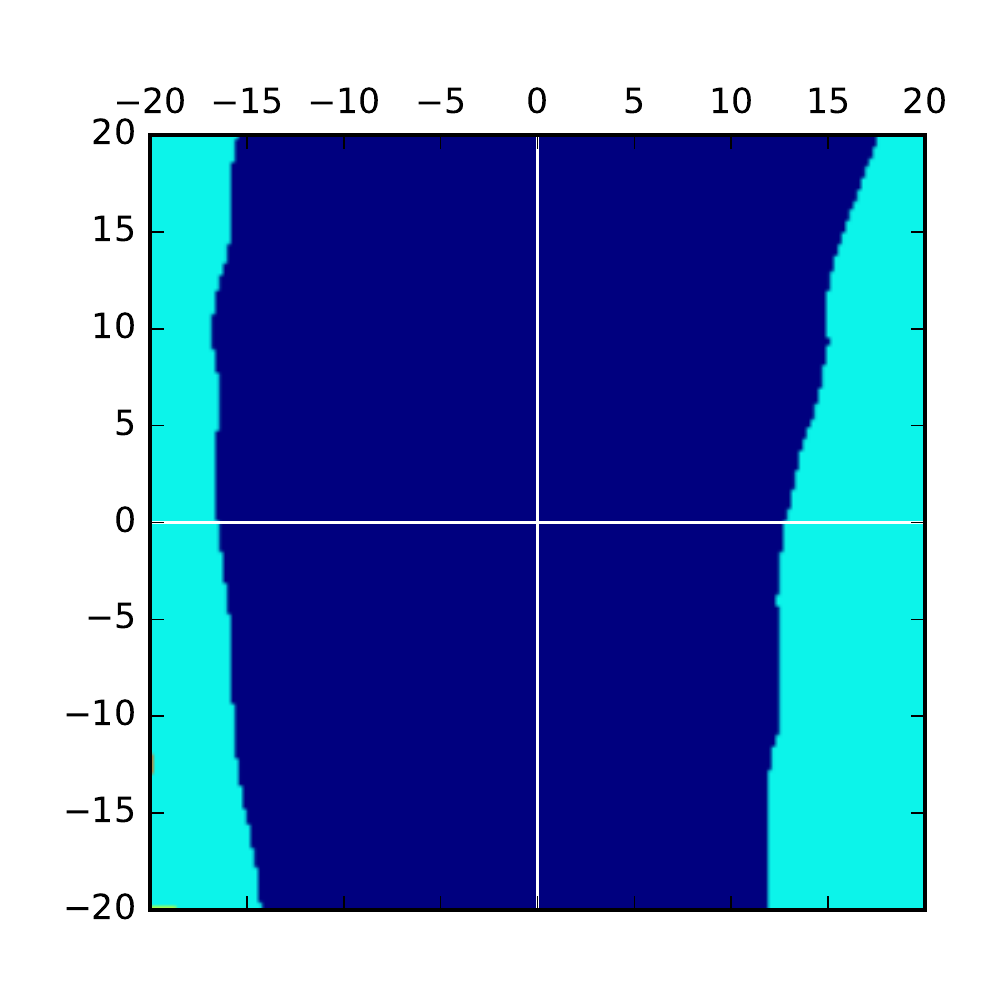} &
  \includegraphics[scale=0.28]{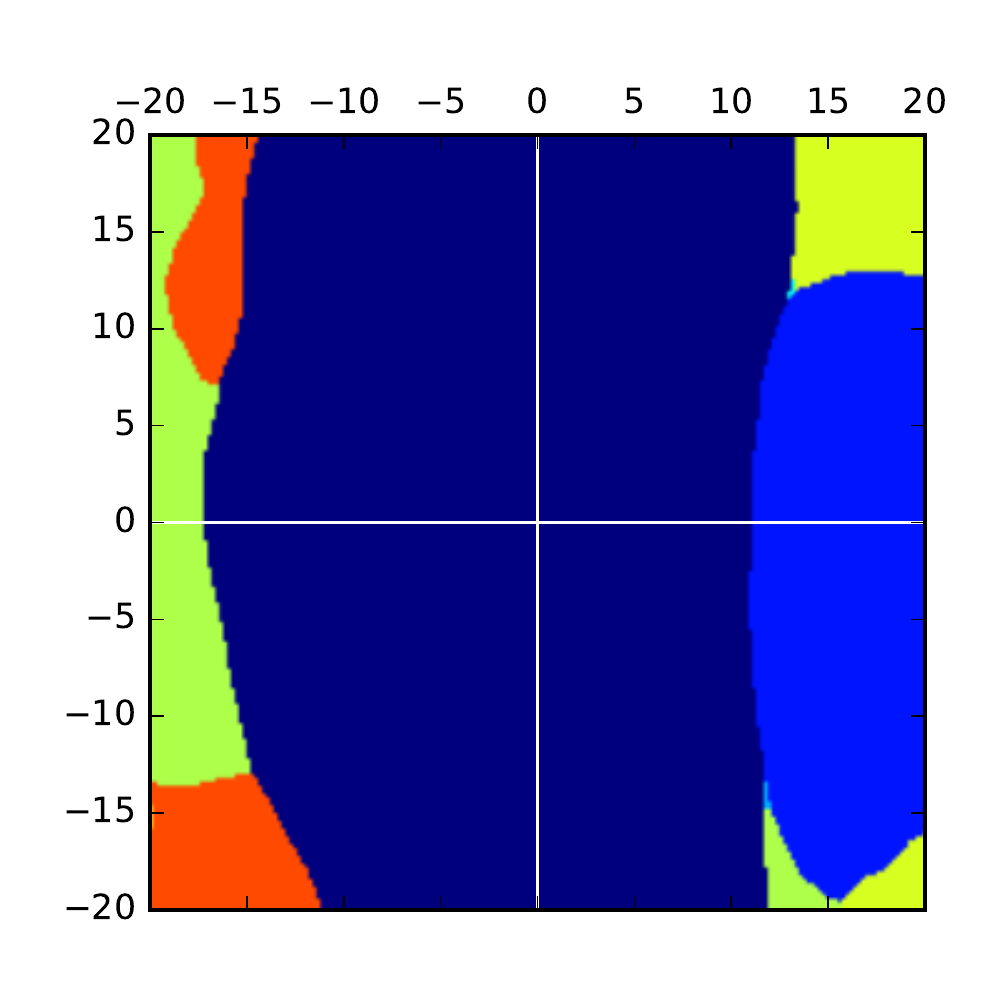}\\
\ifshort
\else
\rot{~~~~~~~~~Zoom-out} &
  \includegraphics[scale=0.28]{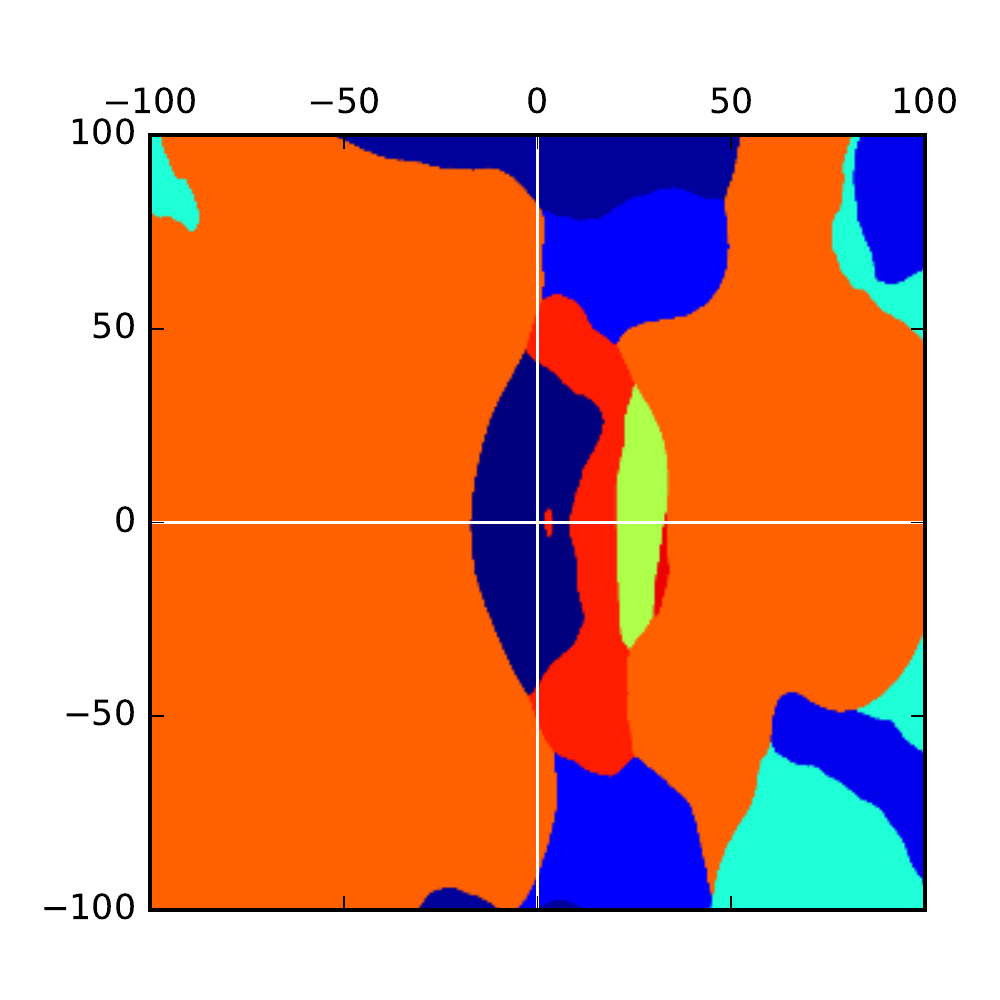} &
  \includegraphics[scale=0.28]{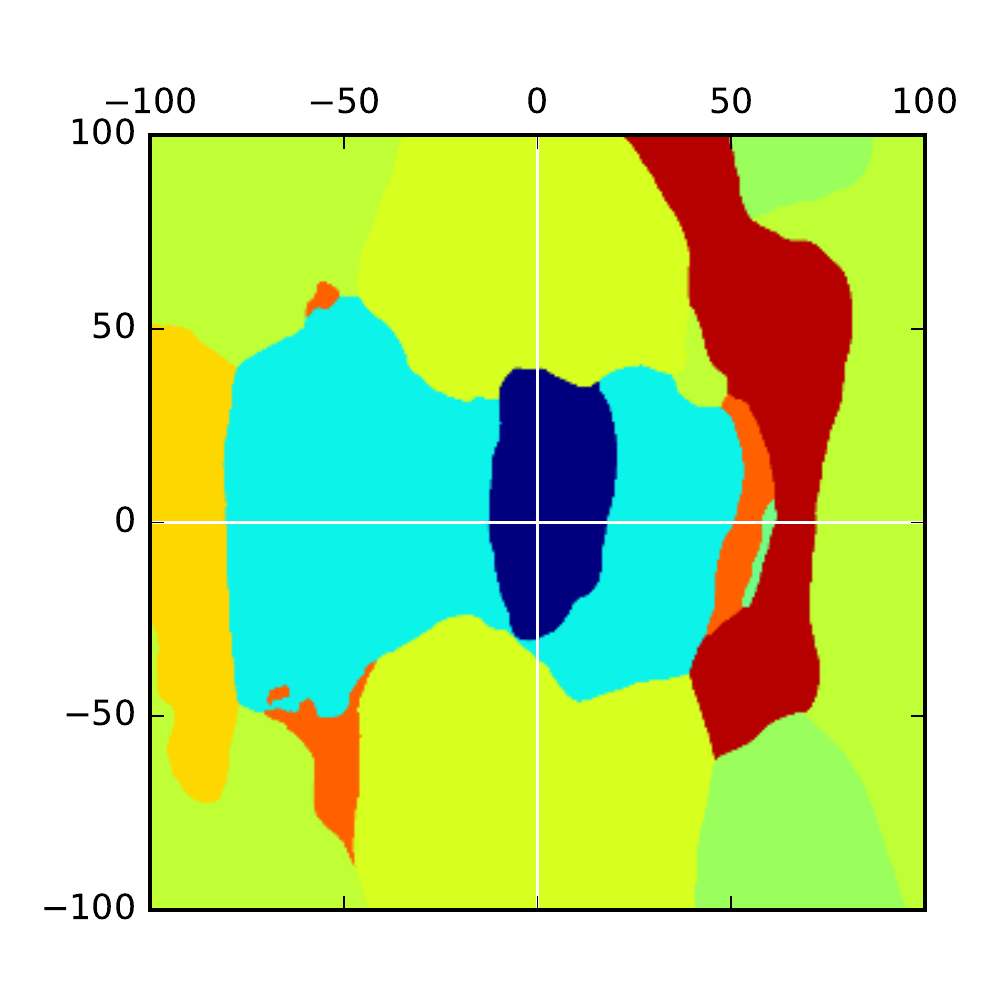} &
  \includegraphics[scale=0.28]{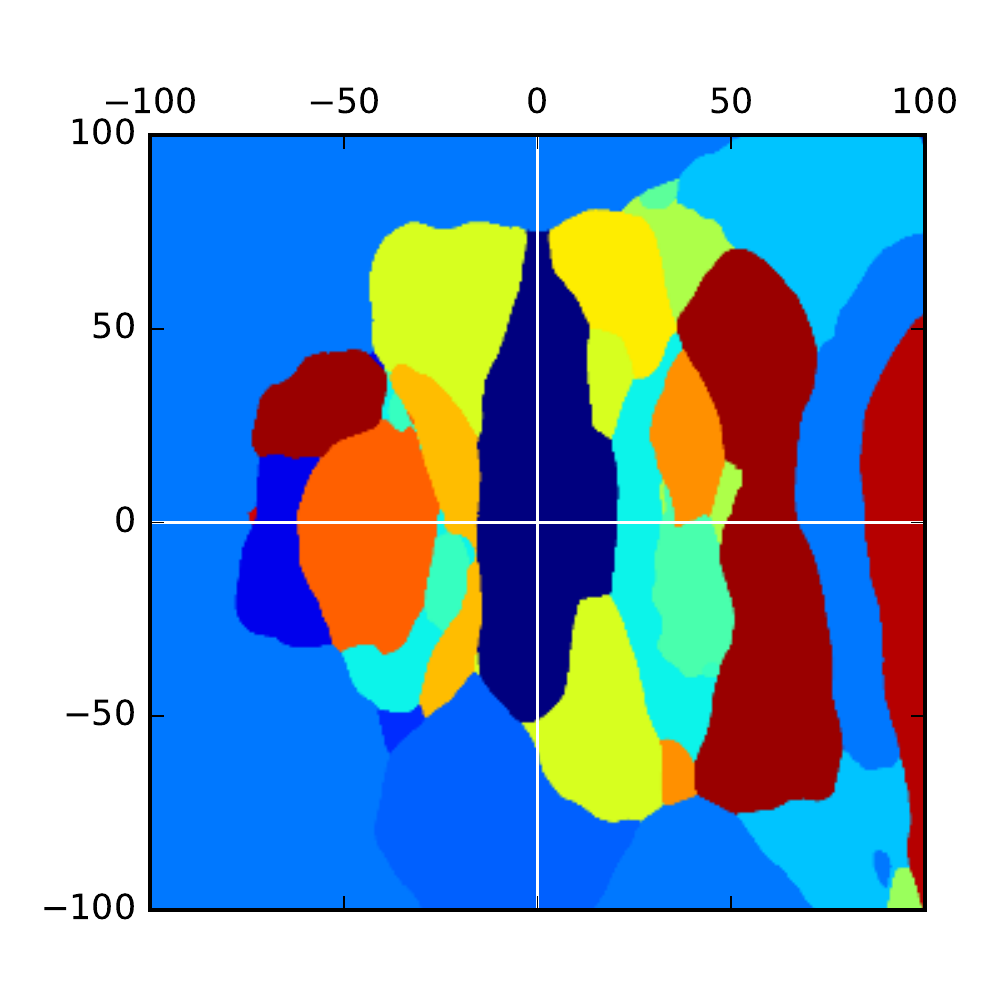} &
  \includegraphics[scale=0.28]{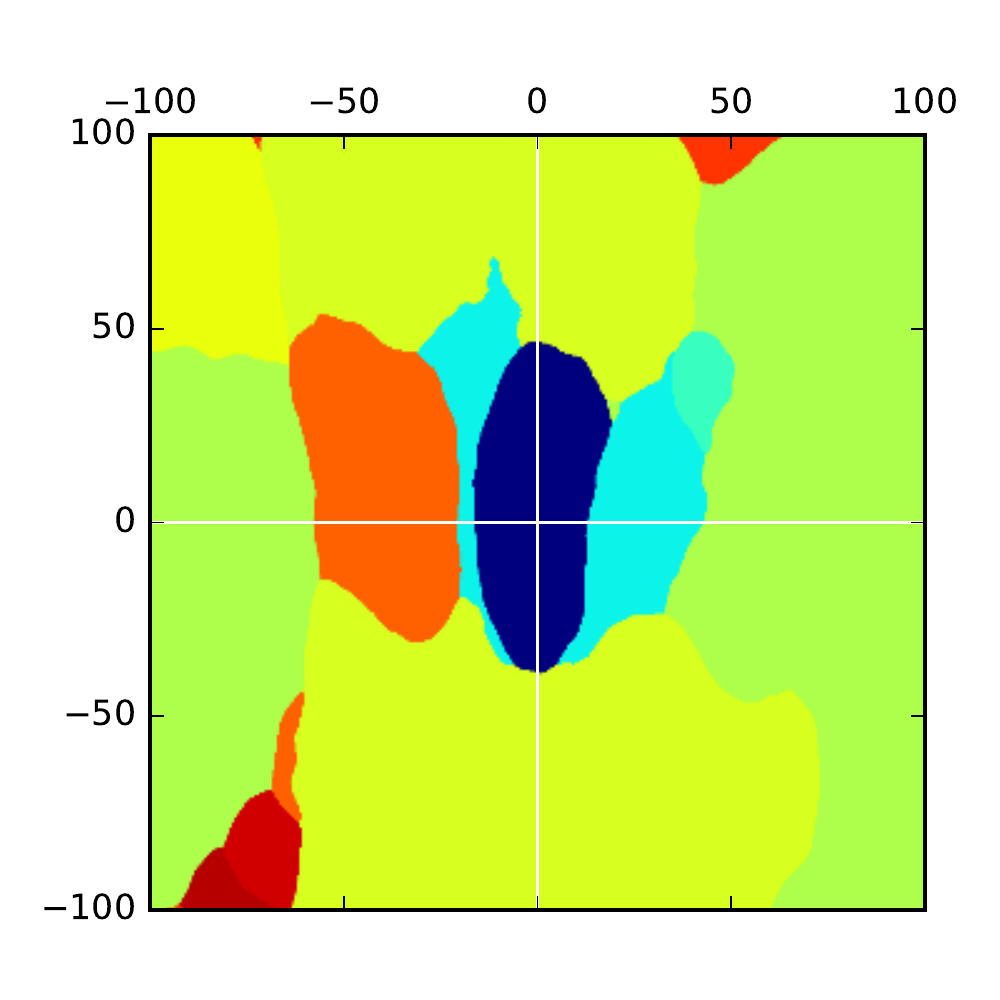} &
  \includegraphics[scale=0.28]{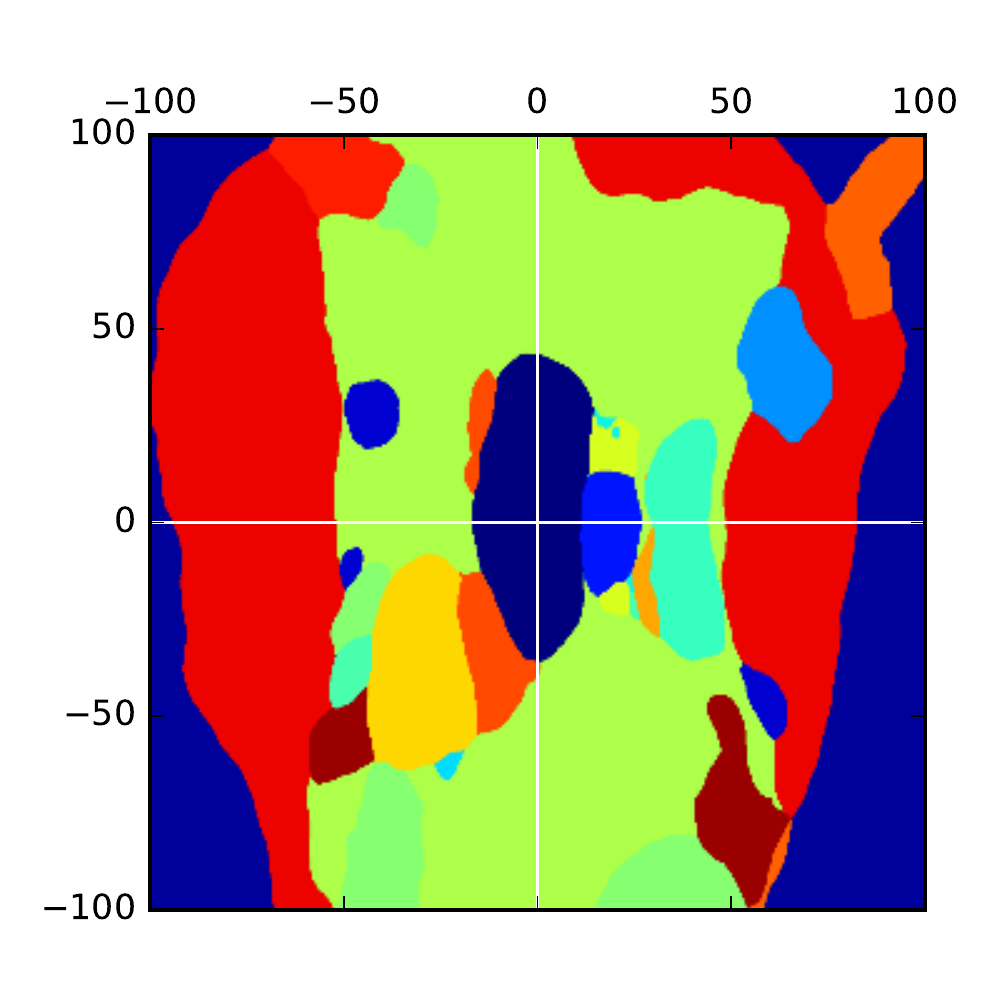}\\
\fi
\end{tabular}
\caption{Decision regions of different models. We pick the same two directions for all plots: one
is the gradient direction of VGG-16 (x-axis), and the other is a random orthogonal direction (y-axis). Each point in the span plane shows the predicted label of the image generated by adding a noise to the original image (e.g., the origin corresponds to the predicted label of the original image). The units of both axises are 1 pixel values. All sub-figure plots the regions
on the span plane using the same color for the same label. The image is
in Figure~\ref{fig:exmp}.}
\label{fig:decision}
\end{figure}

\begin{table}[t]
\centering
\begin{tabular}{|c|c|c|c|c|c|}
\hline
Model & VGG-16 & ResNet-50 & ResNet-101 & ResNet-152 & GoogLeNet\\
\hline
\# of labels &   10 & 9 & 21 & 10 & 21\\
\hline
\end{tabular}
\caption{\small The number of all possible predicted labels for each model in the same plane described in Figure~\ref{fig:decision}.\vspace{-1em}}
\label{tab:num-regions}
\end{table}

\paragraph{Decision boundaries of the non-targeted approaches using a single model.}

We study the decision boundary of different models to understand why
adversarial examples transfer. We choose two normalized orthogonal
directions $\delta_1,\delta_2$, one being the gradient direction of VGG-16 and the other being randomly chosen.
Each point $(u, v)$ in this 2-D plane corresponds to the image
$x+u\delta_1+v\delta_2$, where $x$ is the pixel value vector of the original image. For each model, we plot the
label of the image corresponding to each point, and get Figure~\ref{fig:decision} using the image in Figure~\ref{fig:exmp}.

We can observe that for all models, the region that each model can
predict the image correctly is limited to the central area. Also,
along the gradient direction, the classifiers are soon misled.
One interesting finding is that along this gradient direction,
the first misclassified label for the three ResNet models (corresponding to the light green region) is the label ``orange".
A more detailed study can be found in the appendix (Table~\ref{tab:same-mistake}, Table~\ref{tab:same-mistake-fg} and~\ref{tab:same-mistake-fgs}).
When we look at the zoom-out figures, however,
the labels of images that are far away from the original one are
different for different models, even among ResNet models.

On the other hand, in Table~\ref{tab:num-regions}, we show the
total number of regions in each plane. In fact, for each plane,
there are at most 21 different regions in all planes. Compared with the
1,000 total categories in ImageNet, this is only 2.1\% of all
categories. That means, for all other 97.9\% labels, no targeted adversarial
example exists in each plane. Such a
phenomenon partially explains why fast gradient-based approaches
can hardly find targeted adversarial images.

\begin{figure}[t]
\centering
\begin{minipage}{0.4\linewidth}
\centering
  \includegraphics[scale=0.18]{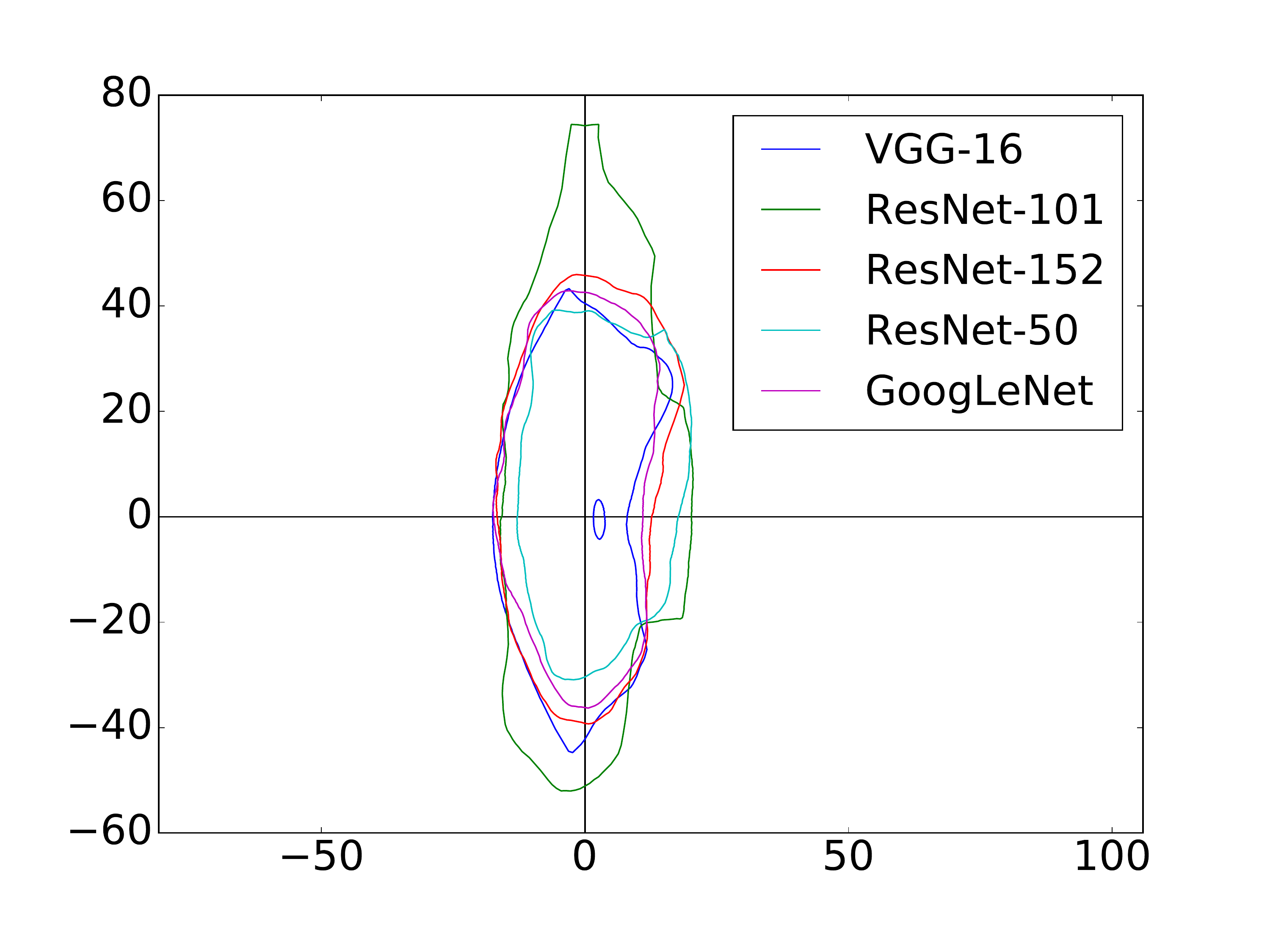}
  \caption{The decision boundary to separate the region within which all points are classified as the ground truth label (encircled by each closed curve) from others. The plane is the same one described in Figure~\ref{fig:decision}. The origin of the coordinate plane corresponds to the original image. The units of both axises are 1 pixel values. }
  \label{fig:nontargeted-boundary}
\end{minipage}
\quad
\begin{minipage}{0.5\linewidth}
\centering
  \includegraphics[scale=0.18]{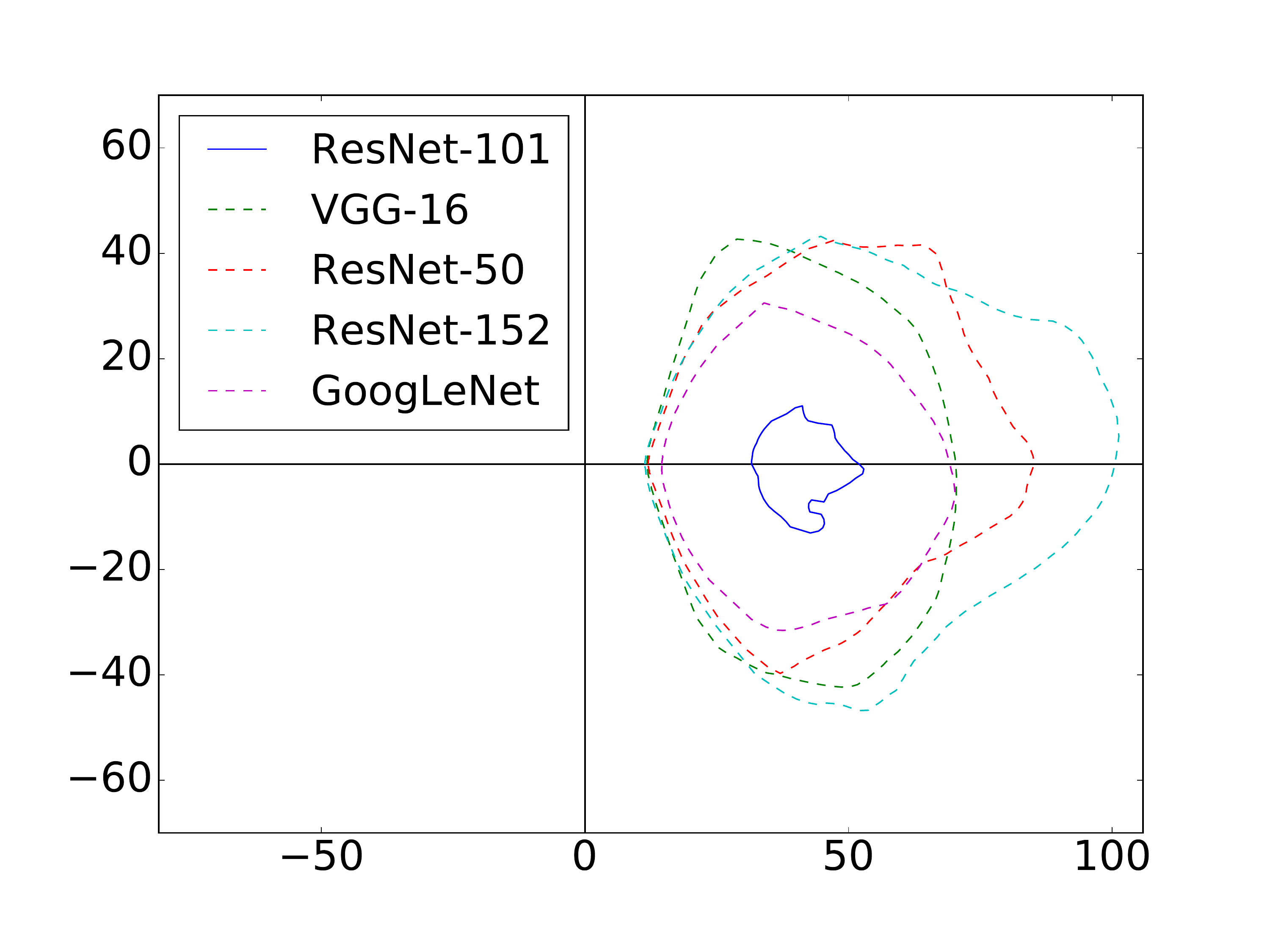}
  \caption{The decision boundary to separate the region within which all points are classified as the target label (encircled by each closed curve) from others. The plane is spanned by the targeted adversarial direction and a random orthogonal direction.
  The targeted adversarial direction is computed as the difference between the original image in Figure~\ref{fig:exmp} and the adversarial 
  image generated by the optimization-based approach for an ensemble. 
  The ensemble contains all models except ResNet-101. The origin of the coordinate plane corresponds to the original image. The units of both axises are 1 pixel values.}
  \label{fig:targeted-boundary}
\end{minipage}
\end{figure}

Further, in Figure~\ref{fig:nontargeted-boundary}, we draw the decision boundaries of all models on the same plane as described above.
We can observe that
\begin{itemize}
\item The boundaries align with each other very well. This
    partially explains why non-targeted adversarial images can transfer
    among models.
\item The boundary diameters along the gradient
    direction is less than the ones along the random direction.
    A potential reason is that
    moving a variable along its gradient direction can change the
    loss function (i.e., the probability of the ground truth label)
    significantly. Therefore along the gradient direction
    it will take fewer steps to move out of the ground truth region than
    a random direction.
\item An interesting finding 
	is that even though we move left along the x-axis, which is equivalent to
    maximizing the ground truth's prediction probability, it also
    reaches the boundary much sooner than moving along a random direction.
    We attribute this to the non-linearity of the loss function: when
    the distortion is larger, the gradient direction also changes
    dramatically. In this case, moving along the
    original gradient direction no longer
    increases the probability to predict the ground truth label (see Figure~\ref{fig:linearity} in the appendix).
\item As for VGG-16 model, there is a small hole within the region corresponding to the ground truth.
    This may partially explain why non-targeted
    adversarial images with small distortion exist, but do not transfer
    well. This hole does not exist in
    other models' decision planes. In this case, non-targeted
    adversarial images in this hole do not transfer.
\end{itemize} 

\paragraph{Decision boundaries of the targeted ensemble-based approaches.}

In addition, we choose the targeted adversarial direction of the ensemble of all models except ResNet-101 and a random orthogonal direction, and we plot decision boundaries on the plane spanned by these two direction vectors in Figure~\ref{fig:targeted-boundary}.
We observe that the regions of images, which are predicted as the target
label, align well for the four models in the ensemble.
However, for the model not used to generate the adversarial image, i.e.,
ResNet-101, it also has a non-empty region such that the prediction is
successfully misled to the target label, although the area is much
smaller. Meanwhile, the region within each closed curve of the models almost has the same center.
\section{Real world example: adversarial examples for \clarifai}
\label{sec:real}

\clarifai is a commercial company providing
state-of-the-art image classification services. We have no 
knowledge about the dataset and types of models used behind 
\clarifai, except that we have black-box access to the services. 
The labels returned from \clarifai are also different from the 
categories in ILSVRC 2012. We submit all 100 original images to \clarifai
and the returned labels are correct based on a subjective measure.

We also submit 400 adversarial images in total, where 200 of them are targeted adversarial examples, and the rest 200 are non-targeted ones. As for the 200 targeted adversarial images, 100 of them are generated using the optimization-based approach based on VGG-16 (the same ones evaluated in Table~\ref{tab:matching-target-opt}), and the rest 100 are generated using the optimization-based approach based on an ensemble of all models except ResNet-152 (the same ones evaluated in Table~\ref{tab:t-trans-label}). The 200 non-targeted adversarial examples are generated similarly (the same ones evaluated in Table~\ref{tab:non-targeted-opt} and~\ref{tab:non-targeted-ensemble-32}).

For non-targeted adversarial examples, we observe that for both the ones generated using VGG-16 and those generated using the ensemble, most of them can transfer to \clarifai.

More importantly, a large proportion of our targeted adversarial examples are misclassified by \clarifai as well. We observe that $57\%$ of the targeted adversarial examples generated using VGG-16, and $76\%$ of the ones generated using the ensemble can mislead \clarifai to predict labels irrelevant to the ground truth.

Further, our experiment shows that for targeted adversarial examples, $18\%$ of those generated using the ensemble model can be predicted as labels close to the target label by \clarifai. The corresponding number for the targeted adversarial examples generated using VGG-16
is $2\%$.  Considering that in the case of attacking \clarifai, the labels given by the target model are different from those given by our models, it is fairly surprising to see that when using the ensemble-based approach, there is still a considerable proportion of our targeted adversarial examples that can mislead this black-box model to make predictions semantically similar to our target labels. All these numbers are computed based
on a subjective measure, and we include some examples in Table~\ref{tab:real-exmp}. More examples can be found in the appendix~(Table~\ref{tab:real-app}).

\begin{longtable}{|C{2cm}|C{1.25cm}|C{2.25cm}|C{1.5cm}|C{2cm}|C{3cm}|}\hline
original image & true label & \clarifai results of original image & target label & targeted adversarial example & \clarifai results of targeted adversarial example\\ \hline

\includegraphics[width=20mm, height=20mm]{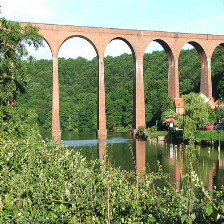} & viaduct & bridge, \newline sight, \newline arch, \newline river, \newline sky & window screen & \includegraphics[width=20mm, height=20mm]{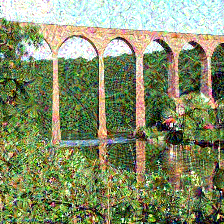} & window, \newline wall, \newline old, \newline decoration, \newline design\\ \hline

\includegraphics[width=20mm, height=20mm]{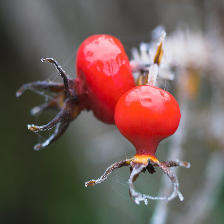} & hip, rose hip, rosehip & fruit, \newline fall, \newline food, \newline little, \newline wildlife & stupa, tope & \includegraphics[width=20mm, height=20mm]{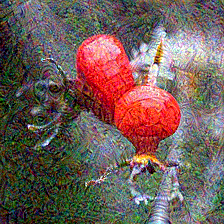} & Buddha, \newline gold, \newline temple, \newline celebration, \newline artistic\\ \hline

\includegraphics[width=20mm, height=20mm]{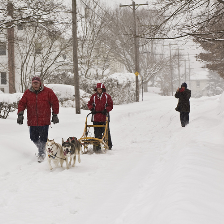} & dogsled, dog sled, dog sleigh & group together, \newline four, \newline sledge, \newline sled, \newline enjoyment & hip, rose hip, rosehip & \includegraphics[width=20mm, height=20mm]{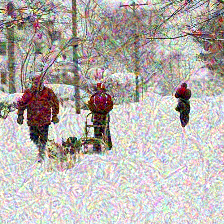} & cherry, \newline branch, \newline fruit, \newline food, \newline season\\ \hline

\includegraphics[width=20mm, height=20mm]{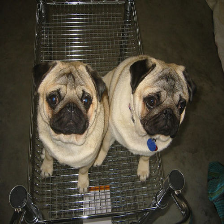} & pug, pug-dog & pug, \newline friendship, \newline adorable, \newline purebred, \newline sit & sea lion & \includegraphics[width=20mm, height=20mm]{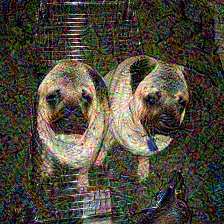} & sea seal, \newline ocean, \newline head, \newline sea, \newline cute\\ \hline

\includegraphics[width=20mm, height=20mm]{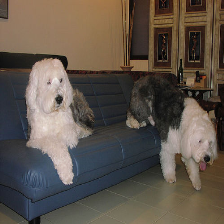} & Old English sheepdog, bobtail & poodle, \newline retriever, \newline loyalty, \newline sit, \newline two & abaya & \includegraphics[width=20mm, height=20mm]{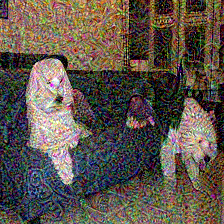} & veil, \newline spirituality, \newline religion, \newline people, \newline illustration\\ \hline

\includegraphics[width=20mm, height=20mm]{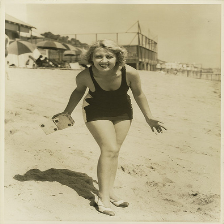} & maillot, tank suit & beach, \newline woman, \newline adult, \newline wear, \newline portrait & amphibian, amphibious vehicle & \includegraphics[width=20mm, height=20mm]{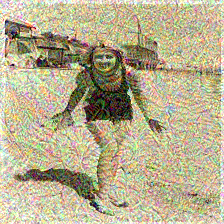} & transportation system, \newline vehicle, \newline man, \newline print, \newline retro\\ \hline

\includegraphics[width=20mm, height=20mm]{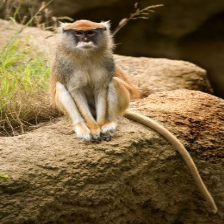} & patas, hussar monkey, Erythrocebus patas & primate, \newline monkey, \newline safari, \newline sit, \newline looking & bee eater & \includegraphics[width=20mm, height=20mm]{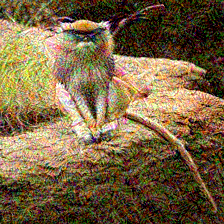} & ornithology, \newline avian, \newline beak, \newline wing, \newline feather\\ \hline

\caption{Original images and adversarial images evaluated over \clarifai. For labels returned from \clarifai, we sort the labels firstly by rareness: how many times a label appears in the \clarifai results for all adversarial images and original images, and secondly by confidence. Only top 5 labels are provided.}
\label{tab:real-exmp}
\end{longtable}

\section{Conclusion}
\label{sec:conc}

In this work, we are the first to conduct an extensive study of the transferability of both non-targeted and targeted adversarial examples generated using different approaches over large models and a large scale dataset. Our results confirm that the transferability for non-targeted adversarial examples are prominent even for large models and a large scale dataset. 
On the other hand, we find that it is hard to use existing approaches to generate targeted adversarial examples whose target labels can transfer. 
We develop novel ensemble-based approaches, and demonstrate that they can generate transferable targeted adversarial examples with
a high success rate. Meanwhile, these new approaches exhibit better performance on
generating non-targeted transferable adversarial examples than previous work.
We also show that both non-targeted and targeted adversarial examples generated using our new approaches can successfully attack \clarifai, which is a black-box image classification system. Furthermore, we study some geometric properties to better understand the transferable adversarial examples.

\subsubsection*{Acknowledgments}

This material is in part based upon work supported by the National 
Science Foundation under Grant No. TWC-1409915.
Any opinions, findings, and conclusions or recommendations expressed
in this material are those of the author(s) and do not necessarily
reflect the views of the National Science Foundation.

\bibliography{iclr2017_conference}
\bibliographystyle{iclr2017_conference}

\newpage

\appendix

\section*{Appendix}

\begin{table}[h]
	\centering
\begin{tabular}{|c|c|c|c|c|c|c|}
\hline
&\small \small ResNet-50&\small ResNet-101&\small ResNet-152&\small GoogLeNet& \small VGG-16\\
\hline
\small Top-1 accuracy&72.5\%&73.8\%&74.6\%&68.7\%&68.3\%\\ \hline
\small Top-5 accuracy&91.0\%&91.7\%&92.1\%&89.0\%&88.3\%\\ \hline
\end{tabular}
\caption{Top-1 and top-5 accuracy of the studied models over the ILSVRC 2012 validation dataset.\vspace{-1.5em}}
\label{tab:accuracy}
\end{table}

\begin{table}
\centering
\begin{tabular}{|c|c|c|c|c|c|c|}
\hline
                  & RMSD  & \small ResNet-152 & \small ResNet-101 & \small ResNet-50 & \small VGG-16 & \small GoogLeNet \\ \hline
\small ResNet-152 & 22.83 & 7\%               & 43\%              & 43\%             & 39\%          & 31\%             \\ \hline
\small ResNet-101 & 23.81 & 40\%              & 6\%               & 41\%             & 42\%          & 34\%             \\ \hline
\small ResNet-50  & 22.86 & 48\%              & 44\%              & 3\%              & 42\%          & 32\%             \\ \hline
\small VGG-16     & 22.51 & 36\%              & 33\%              & 33\%             & 0\%           & 15\%             \\ \hline
\small GoogLeNet  & 22.58 & 66\%              & 71\%              & 62\%             & 49\%          & 2\%              \\ \hline
\end{tabular}
\caption{
Top-5 accuracy of Table~\ref{tab:non-targeted-opt} Panel A. 
Transferability between pairs of models using non-targeted optimization-based approach with a learning rate of 4.
The first column indicates the average RMSD of all adversarial images generated for the model in the corresponding row.
The cell $(i, j)$ indicates the top-5 accuracy of the
adversarial images generated for model $i$ (row) evaluated over model $j$ (column).
Lower value indicates better transferability.
}
\label{tab:top5-non-targeted-opt}
\end{table}

\paragraph{An alternative optimization-based approach to generate adversarial examples.}
An alternative method to generate non-targeted adversarial
examples with large distortion is to revise the optimization objective to
incorporate this distortion constraint.
For example, for non-targeted adversarial image searching, we can optimize
for the following objective.
\[\mathbf{argmin}_{x^\star} -\log{(1-\mathbf{1}_y \cdot J_\theta(x^\star))} + \lambda_1 \mathrm{ReLU}(\tau - d(x, x^\star)) + \lambda_2 \mathrm{ReLU}(d(x, x^\star) - \tau)\]
Optimizing for the above objective has the following three effects:
(1) minimizing $-\log{(1-\mathbf{1}_y \cdot J_\theta(x^\star))}$;
(2) Penalizing the solution if $d(x, x^\star)$ is no more than a threshold $\tau$ (too low); and
(3) Penalizing the solution if $d(x, x^\star)$ is too high.

In our preliminary evaluation, we found that the solutions computed from
the two approaches have similar transferability. We thus omit the results
for this alternative approach.

\paragraph{Transferable non-targeted adversarial images are classified as the same wrong labels.}
Previous work~\cite{fast-gradient-sign} reported the phenomenon that when
evaluating the adversarial images,
different models tend to make the same wrong predictions. This conclusion was
only examined over datasets with 10 categories. In our evaluation,
however, we observe the same phenomenon, albeit we have 1000 possible categories.
We refer to this effect as \emph{the same mistake effect}.

Table~\ref{tab:same-mistake} presents the results based on the adversarial examples generated for VGG-16.
For each pair of models, among all adversarial examples
that both models make wrong predictions,
we compute the percentage that both models make the same mistake.
These percentage numbers are from 12\% to 40\%, which are surprisingly high,
since there are 999 possible categories to be misclassified into 
(see Table~\ref{tab:label-distribution-non-targeted-VGG16} in the appendix for the wrong predicted label distribution of these adversarial examples).
Later in Section~\ref{sec:geo}, we try to explain this phenomenon using decision boundaries.

\begin{table}
\centering
\begin{tabular}{c}
\begin{tabular}{|c|c|c|c|c|c|}
\hline
                  & \small ResNet-152 & \small ResNet-101 & \small ResNet-50 & \small VGG-16 & \small GoogLeNet 
                  \\ \hline
\small ResNet-152 & 100.00\%         & $-$             & $-$            & $-$         & $-$ 
\\ \hline
\small ResNet-101 & 28.57\%             & 100.00\%            & $-$            & $-$         & $-$  
\\ \hline
\small ResNet-50  & 29.87\%             & 40.00\%             & 100.00\%          & $-$         & $-$   
\\ \hline
\small VGG-16     & 19.23\%             & 18.07\%             & 22.89\%            & 100.00\%   & $-$    
\\ \hline
\small GoogLeNet  & 12.82\%             & 20.48\%             & 18.07\%            & 16.84\%         & 100.00\%   
\\ \hline
\end{tabular}
\end{tabular}
\caption{\footnotesize When using the optimization-based approach for VGG-16 model to generate non-targeted adversarial images,
cell $(i, j)$ is the percentage of all transferable adversarial images that are predicted as the same wrong labels
by both models $i$ and $j$ over all adversarial images that are misclassified by both models $i$ and $j$. Notice that this table is symmetric.\vspace{-1em}}
\label{tab:same-mistake}
\end{table}

\begin{figure}
\begin{subfigure}{0.3\linewidth}
\centering
  \includegraphics[scale=0.35]{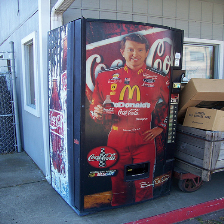}
  \caption{The original image}
  \label{fig:linear-resnet}
\end{subfigure}
\begin{subfigure}{0.7\linewidth}
\centering
  \includegraphics[scale=0.4]{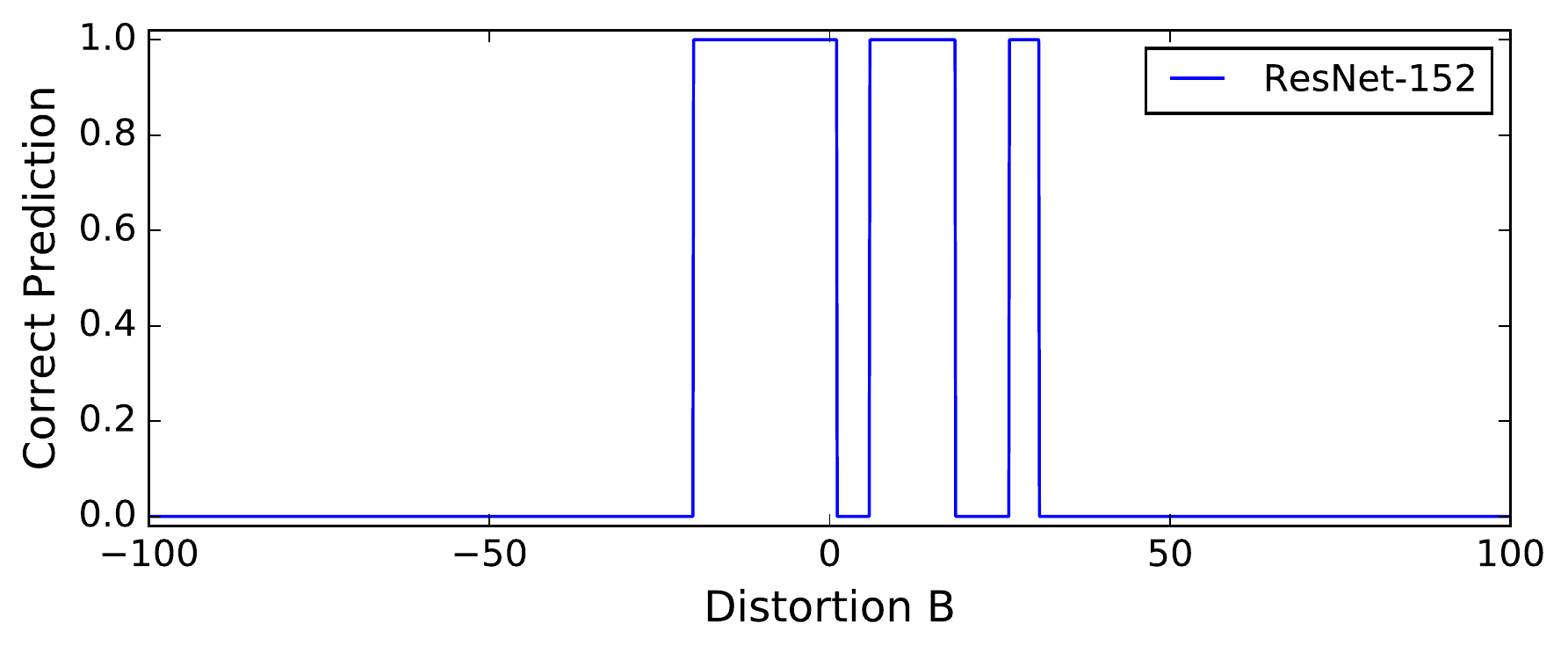}
  \caption{Prediction correctness}
  \label{fig:linear-vgg}
\end{subfigure}
\caption{We add $B\delta$ onto the original image in
	Figure~\ref{fig:linear-resnet}, where $\delta$ is the normalized gradient
	of the non-targeted objective (\ref{obj:non-targeted}) for VGG-16.
	When $x_B$ is evaluated on ResNet-152, we plot 1 if the prediction matches the ground
	truth, or 0 otherwise.\vspace{-1.5em}}
\label{fig:non-contiguous}
\end{figure}

\paragraph{Adversarial images may come from multiple intervals along the
gradient direction.}
In Figure~\ref{fig:non-contiguous}, we show that along the gradient direction
of the non-targeted objective~(\ref{obj:non-targeted}) for VGG-16,
when evaluating the adversarial image $x_B$ on ResNet-152,
$x_B$ will soon become adversarial for small $B$.
While $B$ increases, however, the prediction
changes back to be the correct label, then a wrong label,
then the correct label again, and finally wrong labels.
This indicates that along the gradient of this image, the distortions that can cause the corresponding adversarial images to mislead ResNet-152 form four intervals.

\begin{table}[t]
\centering
\begin{tabular}{|c|c|c|c|c|c|c|}
\hline
                  & RMSD  & \small ResNet-152 & \small ResNet-101 & \small ResNet-50 & \small VGG-16 & \small GoogLeNet \\ \hline
\small ResNet-152 & 23.45 & 23\%              & 29\%              & 31\%             & 39\%          & 33\%             \\ \hline
\small ResNet-101 & 23.49 & 35\%              & 20\%              & 33\%             & 43\%          & 28\%             \\ \hline
\small ResNet-50  & 23.49 & 40\%              & 39\%              & 18\%             & 39\%          & 33\%             \\ \hline
\small VGG-16     & 23.73 & 40\%              & 35\%              & 33\%             & 8\%           & 19\%             \\ \hline
\small GoogLeNet  & 23.45 & 53\%              & 46\%              & 37\%             & 38\%          & 7\%              \\ \hline
\end{tabular}
\caption{
Top-5 accuracy of Table~\ref{tab:non-targeted-opt} Panel B. 
Transferability between pairs of models using non-targeted FG.
The first column indicates the average RMSD of all adversarial images generated for the model in the corresponding row.
The cell $(i, j)$ indicates the top-5 accuracy of the
adversarial images generated for model $i$ (row) evaluated over model $j$ (column).
Lower value indicates better transferability.
}
\label{tab:top5-trans-fg}
\end{table}

\paragraph{Comparison with random perturbations.}
For comparison, we evaluate the test accuracy when we add a Gaussian noise to the 100 images in our test set. We vary the standard deviation of the Gaussian noise from 5 to 40 with a step size of 5. For each specific standard deviation, we generate 100 random noises and add them to each image, resulting in 10,000 noisy images in total. Then we evaluate the accuracy of each model on these 10,000 images, and the results are presented in Table~\ref{tab:noise}. Notice that when setting standard deviation to be $25$, the average RMSD is 23.59, which is comparable to that of non-targeted adversarial examples generated by either optimization-based approaches or fast gradient-based approaches. However, each model can still achieve an accuracy more than $66\%$. This shows that adding random noise is not an effective way to generate adversarial examples, hence the ``transferability" of this approach is significantly worse than either optimization-based approaches or fast gradient-based approaches.

\begin{table}
\centering
\begin{tabular}{|c|c|c|c|c|c|c|}
\hline
\small Standard Deviation & \small RMSD & \small ResNet-152	&\small ResNet-101	&\small ResNet-50	&\small VGG-16	&\small GoogLeNet\\
\hline
 5 &	4.91	& 97.41\%	& 98.96\%	& 98.74\%	& 97.47\%	& 99.29\%\\
\hline
10 &	9.72	& 95.53\%	& 96.72\%	& 96.81\%	& 92.13\%	& 95.94\%\\
\hline
15 &	14.44	& 91.19\%	& 94.22\%	& 92.16\%	& 87.86\%	& 88.50\%\\
\hline
20 &	19.07	& 86.56\%	& 90.38\%	& 84.07\%	& 82.30\%	& 77.84\%\\
\hline
25 &	23.59	& 83.10\%	& 85.53\%	& 78.33\%	& 73.57\%	& 66.84\%\\
\hline
30 &	28.01	& 78.95\%	& 79.04\%	& 71.66\%	& 65.33\%	& 54.93\%\\
\hline
35 &	32.32	& 73.60\%	& 70.89\%	& 62.03\%	& 58.55\%	& 45.13\%\\
\hline
40 &	36.52	& 66.53\%	& 63.09\%	& 50.96\%	& 51.85\%	& 35.61\%\\
\hline
\end{tabular}
\caption{Accuracy of images with random perturbation. The first column reports the standard deviation of the Gaussian noise added to each image. The second column reports the average RMSD over all generated images with the respective standard deviation. For each of the rest column $j$, the cell $(i, j)$ reports model $j$'s accuracy of the noisy images when a Gaussian noise with the respective standard deviation specified in row $i$ is added to each image.}
\label{tab:noise}
\end{table}

\begin{table}
\centering
\begin{tabular}{|c|c|c|c|c|c|c|}
\hline
                  & RMSD  & \small ResNet-152 & \small ResNet-101 & \small ResNet-50 & \small VGG-16 & \small GoogLeNet \\ \hline
\small ResNet-152 & 23.13 & 100\%             & 11\%              & 5\%              & 3\%           & 1\%              \\ \hline
\small ResNet-101 & 23.16 & 9\%               & 100\%             & 7\%              & 2\%           & 1\%              \\ \hline
\small ResNet-50  & 23.06 & 10\%              & 9\%               & 100\%            & 2\%           & 3\%              \\ \hline
\small VGG-16     & 23.59 & 3\%               & 5\%               & 5\%              & 100\%         & 4\%              \\ \hline
\small GoogLeNet  & 22.87 & 1\%               & 2\%               & 1\%              & 3\%           & 100\%            \\ \hline
\end{tabular}
\caption{
Top-5 matching rate of Table~\ref{tab:matching-target-opt}.
The adversarial images are generated using the targeted optimization-based approach with a learning rate of  4.
The first column indicates the average RMSD of all adversarial images generated for the model in the corresponding row.
Cell $(i, j)$ indicates that top-5 matching rate of the targeted adversarial images
generated for model $i$ (row) when evaluated on model $j$ (column).
Higher value indicates more successful transferable target labels.
}
\label{tab:top5-matchingrate-opt}
\end{table}

\begin{table}[h]
\centering
\begin{tabular}{|c|c|c|c|c|c|c|}
\hline
                  & RMSD  & \small ResNet-152 & \small ResNet-101 & \small ResNet-50 & \small VGG-16 & \small GoogLeNet \\ \hline
\small -ResNet-152 & 30.68 & 87\%              & 100\%             & 100\%            & 100\%         & 100\%            \\ \hline
\small -ResNet-101 & 30.76 & 100\%             & 88\%              & 100\%            & 100\%         & 100\%            \\ \hline
\small -ResNet-50  & 30.26 & 99\%              & 99\%              & 86\%             & 99\%          & 99\%             \\ \hline
\small -VGG-16     & 31.13 & 100\%             & 100\%             & 100\%            & 51\%          & 100\%            \\ \hline
\small -GoogLeNet  & 29.70 & 100\%             & 100\%             & 100\%            & 100\%         & 32\%             \\ \hline
\end{tabular}
\caption{
The top-5 matching rate of Table~\ref{tab:t-trans-label}. Matching rate of adversarial images generated using targeted optimization-based approach. The first column indicates the average RMSD of all adversarial images generated for the model in the corresponding row.
Cell $(i, j)$ indicates that top-5 matching rate of the targeted adversarial images generated using
the ensemble of the four models except model $i$ (row) is predicted as the target
label by model $j$ (column). In each row, the minus sign ``$-$" indicates that the model of
the row is not used when generating the attacks.
Higher value indicates more successful transferable target labels.
}
\label{tab:top5-matchingrate-ensemble}
\end{table}

\begin{table}[h]
\centering
\begin{tabular}{|c|c|c|c|c|c|c|}
\hline
                  & RMSD  & \small ResNet-152 & \small ResNet-101 & \small ResNet-50 & \small VGG-16 & \small GoogLeNet \\ \hline
\small -ResNet-152 & 17.17 & 13\%              & 4\%               & 4\%              & 0\%           & 3\%              \\ \hline
\small -ResNet-101 & 17.25 & 2\%               & 11\%              & 3\%              & 0\%           & 3\%              \\ \hline
\small -ResNet-50  & 17.25 & 4\%               & 5\%               & 11\%             & 0\%           & 2\%              \\ \hline
\small -VGG-16     & 17.80 & 4\%               & 7\%               & 5\%              & 20\%          & 4\%              \\ \hline
\small -GoogLeNet  & 17.41 & 3\%               & 2\%               & 3\%              & 0\%           & 15\%             \\ \hline
\end{tabular}
\caption{
Top-5 accuracy of Table~\ref{tab:non-targeted-ensemble-32}.
The first column indicates the average RMSD of all adversarial images generated for the model in the corresponding row.
Cell $(i, j)$ indicates that top-5 accuracy of the non-targeted adversarial images generated using
the ensemble of the four models except model $i$ (row) when evaluated over model $j$ (column).
In each row, the minus sign ``$-$" indicates that the model of
the row is not used when generating the attacks.
Lower value indicates better transferability.
}
\label{tab:top5-non-targeted-ensemble-32}
\end{table}

\begin{table}[h]
\centering
\begin{tabular}{|c|c|c|c|c|c|c|}
\hline
                  & RMSD & \small ResNet-152  & \small ResNet-101  & \small ResNet-50  & \small VGG-16  & \small GoogLeNet \\ \hline
\small ResNet-152  & 1.25 & 0\%               & 86\%              & 87\%             & 93\%          & 96\%             \\ \hline
\small ResNet-101  & 1.24 & 84\%              & 0\%               & 93\%             & 95\%          & 100\%            \\ \hline
\small ResNet-50   & 1.21 & 90\%              & 91\%              & 0\%              & 91\%          & 97\%             \\ \hline
\small VGG-16      & 1.55 & 89\%              & 94\%              & 92\%             & 0\%           & 84\%             \\ \hline
\small GoogLeNet  & 1.27 & 94\%              & 97\%              & 98\%             & 91\%          & 0\%              \\ \hline
\end{tabular}
\caption{
Transferability between pairs of models using non-targeted optimization-based approach with a learning rate of 0.125.
The first column indicates the average RMSD of all adversarial images generated for the model in the corresponding row.
The cell $(i, j)$ indicates the top-1 accuracy of the
adversarial images generated for model $i$ (row) evaluated over model $j$ (column).
Lower value indicates better transferability. Results of top-5 accuracy can be found in Table~\ref{tab:top5-nontargeted-o-small}.
}
\label{tab:top1-nontargeted-o-small}
\end{table}

\begin{table}[h]
\centering
\begin{tabular}{|c|c|c|c|c|c|c|}
\hline
                  & RMSD & \small ResNet-152 & \small ResNet-101 & \small ResNet-50 & \small VGG-16 & \small GoogLeNet \\ \hline
\small ResNet-152 & 1.25 & 24\%              & 100\%             & 100\%            & 99\%          & 100\%            \\ \hline
\small ResNet-101 & 1.24 & 100\%             & 22\%              & 99\%             & 100\%         & 100\%            \\ \hline
\small ResNet-50  & 1.21 & 100\%             & 99\%              & 25\%             & 100\%         & 100\%            \\ \hline
\small VGG-16     & 1.55 & 98\%              & 100\%             & 100\%            & 15\%          & 100\%            \\ \hline
\small GoogLeNet  & 1.27 & 100\%             & 100\%             & 100\%            & 100\%         & 18\%             \\ \hline
\end{tabular}
\caption{
Top-5 accuracy of Table~\ref{tab:top1-nontargeted-o-small}. 
Transferability between pairs of models using non-targeted optimization-based approach with a learning rate of 0.125.
The first column indicates the average RMSD of all adversarial images generated for the model in the corresponding row.
The cell $(i, j)$ indicates the top-5 accuracy of the
adversarial images generated for model $i$ (row) evaluated over model $j$ (column).
Lower value indicates better transferability.
}
\label{tab:top5-nontargeted-o-small}
\end{table}

\begin{table}[h]
\centering
\begin{tabular}{|c|c|c|c|c|c|c|}
\hline
                  & RMSD  & \small ResNet-152 & \small ResNet-101 & \small ResNet-50 & \small VGG-16 & \small GoogLeNet \\ \hline
\small ResNet-152 & 31.35 & 98\%              & 1\%               & 2\%              & 2\%           & 0\%              \\ \hline
\small ResNet-101 & 31.11 & 5\%               & 98\%              & 1\%              & 2\%           & 0\%              \\ \hline
\small ResNet-50  & 31.32 & 3\%               & 2\%               & 99\%             & 1\%           & 1\%              \\ \hline
\small VGG-16     & 31.50 & 1\%               & 1\%               & 2\%              & 97\%          & 1\%              \\ \hline
\small GoogLeNet  & 30.67 & 2\%               & 1\%               & 0\%              & 2\%           & 97\%             \\ \hline
\end{tabular}
\caption{
The adversarial images are generated using the targeted optimization-based approach with a larger learning rate.
The first column indicates the average RMSD of all adversarial images generated for the model in the corresponding row.
Cell $(i, j)$ indicates that top-1 matching rate of the targeted adversarial images
generated for model $i$ (row) when evaluated on model $j$ (column).
Higher value indicates more successful transferable target labels.
We used a larger learning rate to achieve larger RMSD. Results of top-5 matching rate can be found in Table~\ref{tab:top5-matchingrate}.
}
\label{tab:matching-target-opt-large}
\end{table}

\begin{table}[h]
\centering
\begin{tabular}{|c|c|c|c|c|c|c|}
\hline
                  & RMSD  & \small ResNet-152 & \small ResNet-101 & \small ResNet-50 & \small VGG-16 & \small GoogLeNet \\ \hline
\small ResNet-152 & 31.35 & 98\%              & 8\%               & 6\%              & 3\%           & 1\%              \\ \hline
\small ResNet-101 & 31.11 & 10\%              & 98\%              & 5\%              & 3\%           & 1\%              \\ \hline
\small ResNet-50  & 31.32 & 6\%               & 7\%               & 99\%             & 5\%           & 1\%              \\ \hline
\small VGG-16     & 31.50 & 4\%               & 4\%               & 6\%              & 97\%          & 4\%              \\ \hline
\small GoogLeNet  & 30.67 & 3\%               & 1\%               & 2\%              & 3\%           & 97\%             \\ \hline
\end{tabular}
\caption{
The top-5 matching rate of Table~\ref{tab:matching-target-opt-large}.
The adversarial images are generated using the targeted optimization-based approach with a larger learning rate.
The first column indicates the average RMSD of all adversarial images generated for the model in the corresponding row.
Cell $(i, j)$ indicates that top-5 matching rate of the targeted adversarial images
generated for model $i$ (row) when evaluated on model $j$ (column).
Higher value indicates more successful transferable target labels.
We used a larger learning rate to achieve larger RMSD.
}
\label{tab:top5-matchingrate}
\end{table}

\begin{table}[h]
\centering
\begin{tabular}{|c|c|c|c|c|c|c|c|}
\hline
                  & RMSD  & \small ResNet-152 & \small ResNet-101 & \small ResNet-50 & \small VGG-16 & \small GoogLeNet \\ \hline
\small ResNet-152 & 23.11 & 12\%              & 27\%              & 25\%             & 22\%          & 15\%             \\ \hline
\small ResNet-101 & 23.11 & 29\%              & 13\%              & 29\%             & 29\%          & 16\%             \\ \hline
\small ResNet-50  & 23.11 & 34\%              & 28\%              & 10\%             & 25\%          & 23\%             \\ \hline
\small VGG-16     & 23.12 & 25\%              & 20\%              & 23\%             & 0\%           & 8\%              \\ \hline
\small GoogLeNet  & 23.11 & 46\%              & 41\%              & 40\%             & 25\%          & 2\%              \\ \hline
\end{tabular}
\caption{
Transferability between pairs of models using non-targeted FGS.
The first column indicates the average RMSD of all adversarial images generated for the model in the corresponding row.
The cell $(i, j)$ indicates the top-1 accuracy of the
adversarial images generated for model $i$ (row) evaluated over model $j$ (column).
Lower value indicates better transferability. Results of top-5 accuracy can be found in Table~\ref{top5-trans-fgs}.
}
\label{tab:trans-fgs}
\end{table}

\begin{table}[h]
\centering
\begin{tabular}{|c|c|c|c|c|c|c|}
\hline
                  & RMSD  & \small ResNet-152 & \small ResNet-101 & \small ResNet-50 & \small VGG-16 & \small GoogLeNet \\ \hline
\small ResNet-152 & 23.11 & 32\%              & 55\%              & 53\%             & 47\%          & 36\%             \\ \hline
\small ResNet-101 & 23.11 & 56\%              & 33\%              & 50\%             & 46\%          & 40\%             \\ \hline
\small ResNet-50  & 23.11 & 59\%              & 53\%              & 29\%             & 47\%          & 38\%             \\ \hline
\small VGG-16     & 23.12 & 42\%              & 39\%              & 41\%             & 5\%           & 21\%             \\ \hline
\small GoogLeNet  & 23.11 & 71\%              & 74\%              & 62\%             & 53\%          & 11\%             \\ \hline
\end{tabular}
\caption{
Top-5 accuracy of Table~\ref{tab:trans-fgs}. 
Transferability between pairs of models using non-targeted FGS.
The first column indicates the average RMSD of all adversarial images generated for the model in the corresponding row.
The cell $(i, j)$ indicates the top-5 accuracy of the
adversarial images generated for model $i$ (row) evaluated over model $j$ (column).
Lower value indicates better transferability.
}
\label{top5-trans-fgs}
\end{table}

\begin{table}[h]
\centering
\begin{tabular}{|c|c|c|c|c|c|c|}
\hline
                 & RMSD  & \small ResNet-152 & \small ResNet-101 & \small ResNet-50 & \small VGG-16 & \small GoogLeNet \\ \hline
\small -ResNet-152 & 17.25 & 23\%             & 12\%             & 11\%            & 1\%          & 7\%              \\ \hline
\small -ResNet-101 & 17.24 & 15\%             & 19\%             & 11\%            & 2\%          & 6\%              \\ \hline
\small -ResNet-50  & 17.24 & 15\%             & 13\%             & 19\%            & 2\%          & 8\%              \\ \hline
\small -VGG-16     & 17.24 & 17\%             & 15\%             & 12\%            & 23\%         & 7\%              \\ \hline
\small -GoogLeNet & 17.24 & 14\%             & 11\%             & 10\%            & 2\%          & 19\%             \\ \hline
\end{tabular}
\caption{
Transferability between pairs of models using non-targeted ensemble FG. The first column indicates the average RMSD of the generated adversarial images.
Cell $(i, j)$ indicates that top-1 accuracy of the non-targeted adversarial images generated using
the ensemble of the four models except model $i$ (row) when evaluated over model $j$ (column).
In each row, the minus sign ``$-$" indicates that the model of
the row is not used when generating the attacks.
Lower value indicates better transferability. Results of top-5 accuracy can be found in Table~\ref{tab:ensemble-fg-top5}.
}
\label{tab:ensemble-fg-top1}
\end{table}

\begin{table}[h]
\centering
\begin{tabular}{|c|c|c|c|c|c|c|}
\hline
                  & RMSD  & \small ResNet-152 & \small ResNet-101 & \small ResNet-50 & \small VGG-16 & \small GoogLeNet \\ \hline
\small -ResNet-152 & 17.25 & 47\%              & 33\%              & 28\%             & 15\%          & 21\%             \\ \hline
\small -ResNet-101 & 17.24 & 37\%              & 42\%              & 30\%             & 14\%          & 26\%             \\ \hline
\small -ResNet-50  & 17.24 & 38\%              & 32\%              & 39\%             & 13\%          & 23\%             \\ \hline
\small -VGG-16     & 17.24 & 37\%              & 36\%              & 37\%             & 44\%          & 32\%             \\ \hline
\small -GoogLeNet  & 17.24 & 32\%              & 30\%              & 28\%             & 13\%          & 46\%             \\ \hline
\end{tabular}
\caption{
Top-5 accuracy of Table~\ref{tab:ensemble-fg-top1}.
Transferability between pairs of models using non-targeted ensemble FG. The first column indicates the average RMSD of the generated adversarial images.
Cell $(i, j)$ indicates that top-5 accuracy of the non-targeted adversarial images generated using
the ensemble of the four models except model $i$ (row) when evaluated over model $j$ (column).
In each row, the minus sign ``$-$" indicates that the model of
the row is not used when generating the attacks.
Lower value indicates better transferability.
}
\label{tab:ensemble-fg-top5}
\end{table}

\begin{table}[h]
\centering
\begin{tabular}{|c|c|c|c|c|c|c|}
\hline
                   & RMSD  & \small ResNet-152 & \small ResNet-101 & \small ResNet-50 & \small VGG-16 & \small GoogLeNet \\ \hline
\small -ResNet-152 & 17.41 & 26\%              & 10\%              & 11\%             & 2\%           & 7\%              \\ \hline
\small -ResNet-101 & 17.40 & 18\%              & 21\%              & 13\%             & 2\%           & 4\%              \\ \hline
\small -ResNet-50  & 17.40 & 15\%              & 13\%              & 20\%             & 4\%           & 8\%              \\ \hline
\small -VGG-16     & 17.40 & 17\%              & 16\%              & 11\%             & 23\%          & 7\%              \\ \hline
\small -GoogLeNet  & 17.40 & 15\%              & 12\%              & 10\%             & 3\%           & 22\%             \\ \hline
\end{tabular}
\caption{
Transferability between pairs of models using non-targeted ensemble FGS. The first column indicates the average RMSD of the generated adversarial images.
Cell $(i, j)$ indicates that top-1 accuracy of the non-targeted adversarial images generated using
the ensemble of the four models except model $i$ (row) when evaluated over model $j$ (column).
In each row, the minus sign ``$-$" indicates that the model of
the row is not used when generating the attacks.
Lower value indicates better transferability. Results of top-5 accuracy can be found in Table~\ref{tab:ensemble-fgs-top5}.
}
\label{tab:ensemble-fgs-top1}
\end{table}

\begin{table}[h]
\centering
\begin{tabular}{|c|c|c|c|c|c|c|}
\hline
                  & RMSD  & \small ResNet-152 & \small ResNet-101 & \small ResNet-50 & \small VGG-16 & \small GoogLeNet \\ \hline
\small -ResNet-152 & 17.41 & 50\%              & 35\%              & 30\%             & 20\%          & 26\%             \\ \hline
\small -ResNet-101 & 17.40 & 38\%              & 48\%              & 33\%             & 19\%          & 26\%             \\ \hline
\small -ResNet-50  & 17.40 & 38\%              & 36\%              & 41\%             & 18\%          & 28\%             \\ \hline
\small -VGG-16     & 17.40 & 37\%              & 36\%              & 36\%             & 48\%          & 33\%             \\ \hline
\small -GoogLeNet  & 17.40 & 35\%              & 29\%              & 31\%             & 19\%          & 53\%             \\ \hline
\end{tabular}
\caption{
Top-5 accuracy of Table~\ref{tab:ensemble-fgs-top1}.
Transferability between pairs of models using non-targeted ensemble FGS. The first column indicates the average RMSD of the generated adversarial images.
Cell $(i, j)$ indicates that top-5 accuracy of the non-targeted adversarial images generated using
the ensemble of the four models except model $i$ (row) when evaluated over model $j$ (column).
In each row, the minus sign ``$-$" indicates that the model of
the row is not used when generating the attacks.
Lower value indicates better transferability.
}
\label{tab:ensemble-fgs-top5}
\end{table}

\begin{table}[h]
\centering
\begin{tabular}{|c|c|c|c|c|c|c|c|}
\hline
                  & RMSD  & \small ResNet-152 & \small ResNet-101 & \small ResNet-50 & \small VGG-16 & \small GoogLeNet \\ \hline
\small ResNet-152 & 23.55 & 1\%               & 2\%               & 0\%              & 0\%           & 1\%              \\ \hline
\small ResNet-101 & 23.56 & 1\%               & 1\%               & 0\%              & 0\%           & 1\%              \\ \hline
\small ResNet-50  & 23.56 & 1\%               & 1\%               & 1\%              & 0\%           & 0\%              \\ \hline
\small VGG-16     & 23.95 & 1\%               & 1\%               & 0\%              & 1\%           & 1\%              \\ \hline
\small GoogLeNet  & 23.63 & 1\%               & 1\%               & 0\%              & 1\%           & 1\%              \\ \hline
\end{tabular}
\caption{
The adversarial images are generated using the targeted FG.
The first column indicates the average RMSD of all adversarial images generated for the model in the corresponding row. The first column indicates the average RMSD of the generated adversarial images. Cell $(i, j)$ indicates that top-1 matching rate of the targeted adversarial images
generated for model $i$ (row) when evaluated on model $j$ (column).
Higher value indicates more successful transferable target labels.
}
\label{tab:fg-targeted}
\end{table}

\begin{table}[h]
\centering
\begin{tabular}{|c|c|c|c|c|c|c|}
\hline
                  & \small ResNet-152 & \small ResNet-101 & \small ResNet-50 & \small VGG-16 & \small GoogLeNet \\ \hline
\small ResNet-152 & 100.00\%             & $-$               & $-$              & $-$           & $-$              \\ \hline
\small ResNet-101 & 33.33\%              & 100.00\%             & $-$              & $-$           & $-$              \\ \hline
\small ResNet-50  & 24.00\%              & 35.00\%              & 100.00\%            & $-$           & $-$              \\ \hline
\small VGG-16     & 17.50\%              & 19.05\%              & 21.18\%             & 100.00\%         & $-$              \\ \hline
\small GoogLeNet  & 15.38\%              & 14.81\%              & 13.10\%             & 15.05\%          & 100.00\%            \\ \hline
\end{tabular}
\caption{
When using non-targeted FG for VGG-16 model to generate adversarial images,
cell $(i, j)$ is the percentage of all transferable adversarial images that are predicted as the same wrong labels
by both models $i$ and $j$ over all adversarial images that are misclassified by both models $i$ and $j$. Notice that the table is symmetric.
}
\label{tab:same-mistake-fg}
\end{table}

\begin{table}[h]
\centering
\begin{tabular}{|c|c|c|c|c|c|c|}
\hline
                  & \small ResNet-152 & \small ResNet-101 & \small ResNet-50 & \small VGG-16 & \small GoogLeNet \\ \hline
\small ResNet-152 & 100.00\%             & $-$               & $-$              & $-$           & $-$              \\ \hline
\small ResNet-101 & 33.78\%              & 100.00\%          & $-$              & $-$           & $-$              \\ \hline
\small ResNet-50  & 36.62\%              & 43.24\%           & 100.00\%         & $-$           & $-$              \\ \hline
\small VGG-16     & 16.00\%              & 20.00\%           & 23.38\%          & 100.00\%      & $-$              \\ \hline
\small GoogLeNet  & 14.86\%              & 20.25\%           & 13.16\%          & 19.57\%       & 100.00\%         \\ \hline
\end{tabular}
\caption{
When using non-targeted FGS for VGG-16 model to generate adversarial images,
cell $(i, j)$ is the percentage of all transferable adversarial images that are predicted as the same wrong labels
by both models $i$ and $j$ over all adversarial images that are misclassified by both models $i$ and $j$. Notice that the table is symmetric.
}
\label{tab:same-mistake-fgs}
\end{table}

\begin{table}[h]
\centering
\begin{tabular}{|c|c|c|c|c|c|c|}
\hline
                  & RMSD  & \small ResNet-152 & \small ResNet-101 & \small ResNet-50 & \small VGG-16 & \small GoogLeNet \\ \hline
\small -ResNet-152 & 31.05 & 1\%               & 1\%               & 0\%              & 1\%           & 1\%              \\ \hline
\small -ResNet-101 & 30.94 & 1\%               & 1\%               & 0\%              & 1\%           & 0\%              \\ \hline
\small -ResNet-50  & 31.12 & 1\%               & 1\%               & 0\%              & 1\%           & 1\%              \\ \hline
\small -VGG-16     & 30.57 & 1\%               & 1\%               & 0\%              & 1\%           & 1\%              \\ \hline
\small -GoogLeNet  & 30.47 & 1\%               & 1\%               & 0\%              & 1\%           & 0\%              \\ \hline
\end{tabular}
\caption{
Transferability between pairs of models using targeted ensemble FG.
The first column indicates the average RMSD of all adversarial images generated for the model in the corresponding row.
Cell $(i, j)$ indicates that top-1 matching rate of the targeted adversarial images generated using
the ensemble of the four models except model $i$ (row) is predicted as the target
label by model $j$ (column). In each row, the minus sign ``$-$" indicates that the model of
the row is not used when generating the attacks.
Higher value indicates more successful transferable target labels. Results of top-5 matching rate can be found in Table~\ref{tab:top5-matchingrate-ensemble-fg}.
}
\label{tab:top1-matchingrate-ensemble-fg}
\end{table}

\begin{table}[h]
\centering
\begin{tabular}{|c|c|c|c|c|c|c|}
\hline
                  & RMSD  & \small ResNet-152 & \small ResNet-101 & \small ResNet-50 & \small VGG-16 & \small GoogLeNet \\ \hline
\small -ResNet-152 & 31.05 & 1\%               & 2\%               & 1\%              & 3\%           & 2\%              \\ \hline
\small -ResNet-101 & 30.94 & 1\%               & 1\%               & 1\%              & 1\%           & 2\%              \\ \hline
\small -ResNet-50  & 31.12 & 1\%               & 2\%               & 1\%              & 3\%           & 2\%              \\ \hline
\small -VGG-16     & 30.57 & 2\%               & 2\%               & 2\%              & 1\%           & 1\%              \\ \hline
\small -GoogLeNet  & 30.47 & 1\%               & 1\%               & 1\%              & 2\%           & 2\%              \\ \hline
\end{tabular}
\caption{
Top-5 matching rate of Table~\ref{tab:top1-matchingrate-ensemble-fg}.
Transferability between pairs of models using targeted ensemble FG.
The first column indicates the average RMSD of all adversarial images generated for the model in the corresponding row.
Cell $(i, j)$ indicates that top-5 matching rate of the targeted adversarial images generated using
the ensemble of the four models except model $i$ (row) is predicted as the target
label by model $j$ (column). In each row, the minus sign ``$-$" indicates that the model of
the row is not used when generating the attacks.
Higher value indicates more successful transferable target labels.
}
\label{tab:top5-matchingrate-ensemble-fg}
\end{table}

\begin{table}[h]
\centering
\begin{tabular}{|c|c|c|c|c|c|c|}
\hline
                  & RMSD  & \small ResNet-152 & \small ResNet-101 & \small ResNet-50 & \small VGG-16 & \small GoogLeNet \\ \hline
\small -ResNet-152 & 30.42 & 1\%               & 1\%               & 1\%              & 1\%           & 1\%              \\ \hline
\small -ResNet-101 & 30.42 & 1\%               & 1\%               & 1\%              & 1\%           & 1\%              \\ \hline
\small -ResNet-50  & 30.42 & 1\%               & 1\%               & 0\%              & 1\%           & 1\%              \\ \hline
\small -VGG-16     & 30.41 & 1\%               & 1\%               & 0\%              & 1\%           & 1\%              \\ \hline
\small -GoogLeNet  & 30.42 & 1\%               & 1\%               & 1\%              & 1\%           & 1\%              \\ \hline
\end{tabular}
\caption{
Transferability between pairs of models using targeted ensemble FGS.
The first column indicates the average RMSD of all adversarial images generated for the model in the corresponding row.
Cell $(i, j)$ indicates that top-1 matching rate of the targeted adversarial images generated using
the ensemble of the four models except model $i$ (row) is predicted as the target
label by model $j$ (column). In each row, the minus sign ``$-$" indicates that the model of
the row is not used when generating the attacks.
Higher value indicates more successful transferable target labels. Results are top-5 matching rate can be found in Table~\ref{tab:top5-matchingrate-ensemble-fgs}.
}
\label{tab:top1-matchingrate-ensemble-fgs}
\end{table}

\begin{table}[h]
\centering
\begin{tabular}{|c|c|c|c|c|c|c|}
\hline
                  & RMSD  & \small ResNet-152 & \small ResNet-101 & \small ResNet-50 & \small VGG-16 & \small GoogLeNet \\ \hline
\small -ResNet-152 & 30.42 & 2\%               & 1\%               & 2\%              & 1\%           & 2\%              \\ \hline
\small -ResNet-101 & 30.42 & 2\%               & 1\%               & 1\%              & 1\%           & 2\%              \\ \hline
\small -ResNet-50  & 30.42 & 2\%               & 1\%               & 1\%              & 1\%           & 2\%              \\ \hline
\small -VGG-16     & 30.41 & 3\%               & 2\%               & 2\%              & 1\%           & 1\%              \\ \hline
\small -GoogLeNet  & 30.42 & 2\%               & 2\%               & 2\%              & 2\%           & 2\%              \\ \hline
\end{tabular}
\caption{
Top-5 matching rate of Table~\ref{tab:top1-matchingrate-ensemble-fgs}.
Transferability between pairs of models using targeted ensemble FGS.
The first column indicates the average RMSD of all adversarial images generated for the model in the corresponding row.
Cell $(i, j)$ indicates that top-5 matching rate of the targeted adversarial images generated using
the ensemble of the four models except model $i$ (row) is predicted as the target
label by model $j$ (column). In each row, the minus sign ``$-$" indicates that the model of
the row is not used when generating the attacks.
Higher value indicates more successful transferable target labels.
}
\label{tab:top5-matchingrate-ensemble-fgs}
\end{table}

\begin{table}[h]
\centering
\begin{tabular}{|c|c|c|c|c|}
\hline
\small ResNet-152         & \small ResNet-101          & \small ResNet-50              & \small VGG-16                 & \small GoogleNet              \\ \hline
\small jigsaw puzzle(8\%) & \small jigsaw puzzle(12\%) & \small jigsaw puzzle(12\%)    & \small jigsaw puzzle(15\%)    & \small prayer rug(12\%)       \\ \hline
\small acorn(3\%)         & \small starfish(3\%)       & \small African chameleon(2\%) & \small African chameleon(7\%) & \small jigsaw puzzle(7\%)     \\ \hline
\small lycaenid(2\%)      & \small strawberry(2\%)     & \small strawberry(2\%)        & \small prayer rug(5\%)        & \small stole(6\%)             \\ \hline
\small ram(2\%)           & \small wild boar(2\%)      & \small starfish(2\%)          & \small apron(4\%)             & \small African chameleon(4\%) \\ \hline
\small maze(2\%)          & \small dishrag(2\%)        & \small greenhouse(2\%)        & \small sarong(3\%)            & \small mitten(3\%)            \\ \hline
\end{tabular}
\caption{When using non-targeted optimization-based approach for VGG-16 model to generate adversarial images,
 column $i$ indicates the top 5 common incorrect labels predicted by model $i$.
 The value in the parentheses is the percentage of the predicted label.}
 \label{tab:label-distribution-non-targeted-VGG16}
\end{table}

\begin{table}[h]
\centering
\begin{tabular}{|c|c|c|c|c|c|c|}
\hline
                  & \small ResNet-152 & \small ResNet-101 & \small ResNet-50 & \small VGG-16 & \small GoogLeNet 
                  \\ \hline
\small ResNet-152  & 1.00            & $-$           & $-$          & $-$       & $-$          
\\ \hline
\small ResNet-101  & 0.04           & 1.00           & $-$          & $-$       & $-$          
\\ \hline
\small ResNet-50   & 0.03           & 0.03           & 1.00           & $-$       & $-$          
\\ \hline
\small VGG-16      & 0.02           & 0.02           & 0.02          & 1.00       & $-$          
\\ \hline
\small GoogLeNet  & 0.01           & 0.01           & 0.01          & 0.02       & 1.00          
\\ \hline
\end{tabular}
	\caption{Average cosine value of the angle between gradient directions of two models. Notice that the dot-product of two normalized vectors is the cosine value of the angle between them, for each image, we compute the dot-product of normalized gradient directions with respect to model $i$ (row) and model $j$ (column), and the value in cell $(i, j)$ is the average over dot-product values of all images. Notice that this table is symmetric.}
	\label{tab:cosine-between-models}
\end{table}


\begin{figure}[b]
\begin{subfigure}[b]{0.32\textwidth}
\centering
  \includegraphics[scale=0.2]{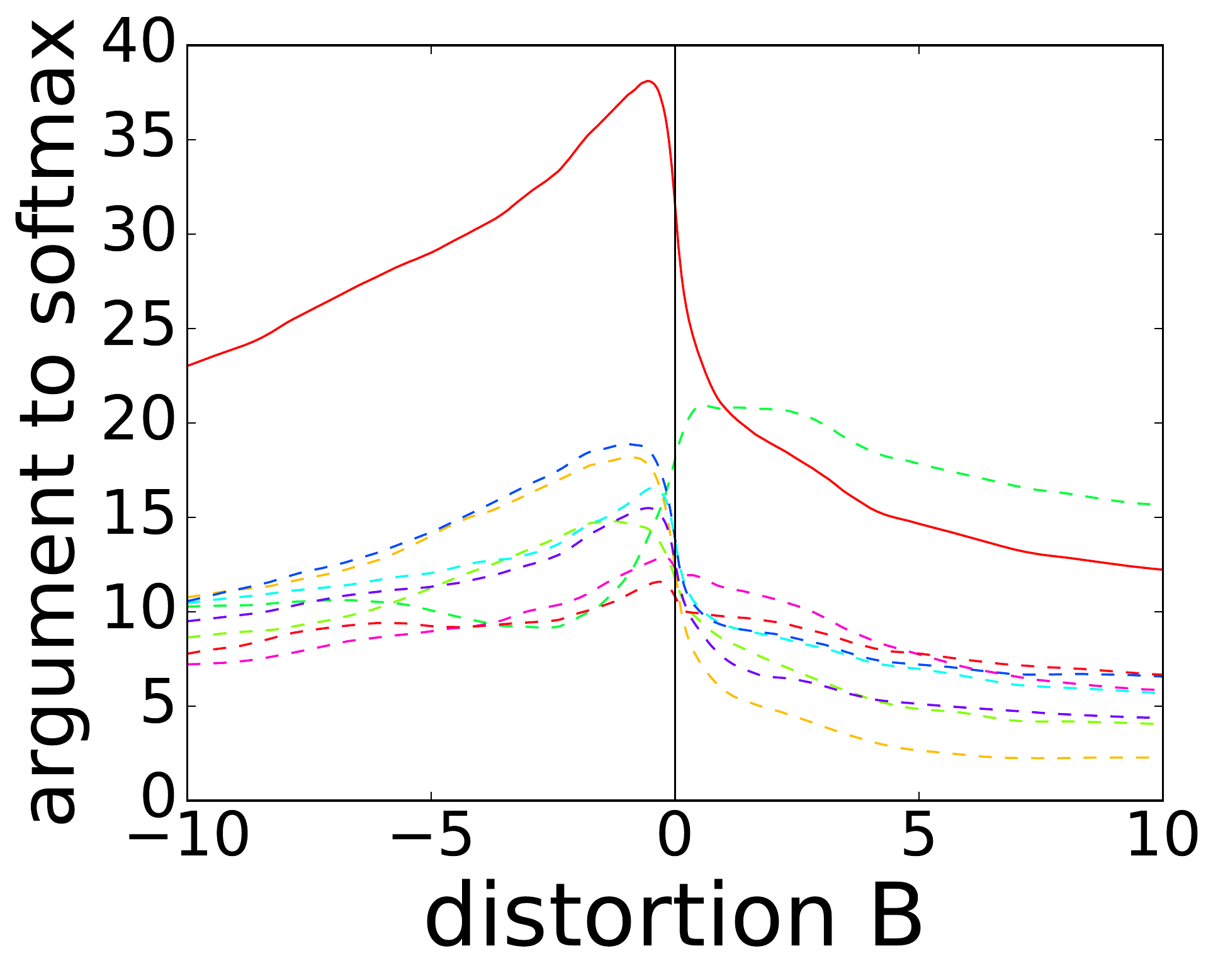}
  \caption{GoogLeNet}
\end{subfigure}
\begin{subfigure}[b]{0.34\textwidth}
\centering
  \includegraphics[scale=0.2]{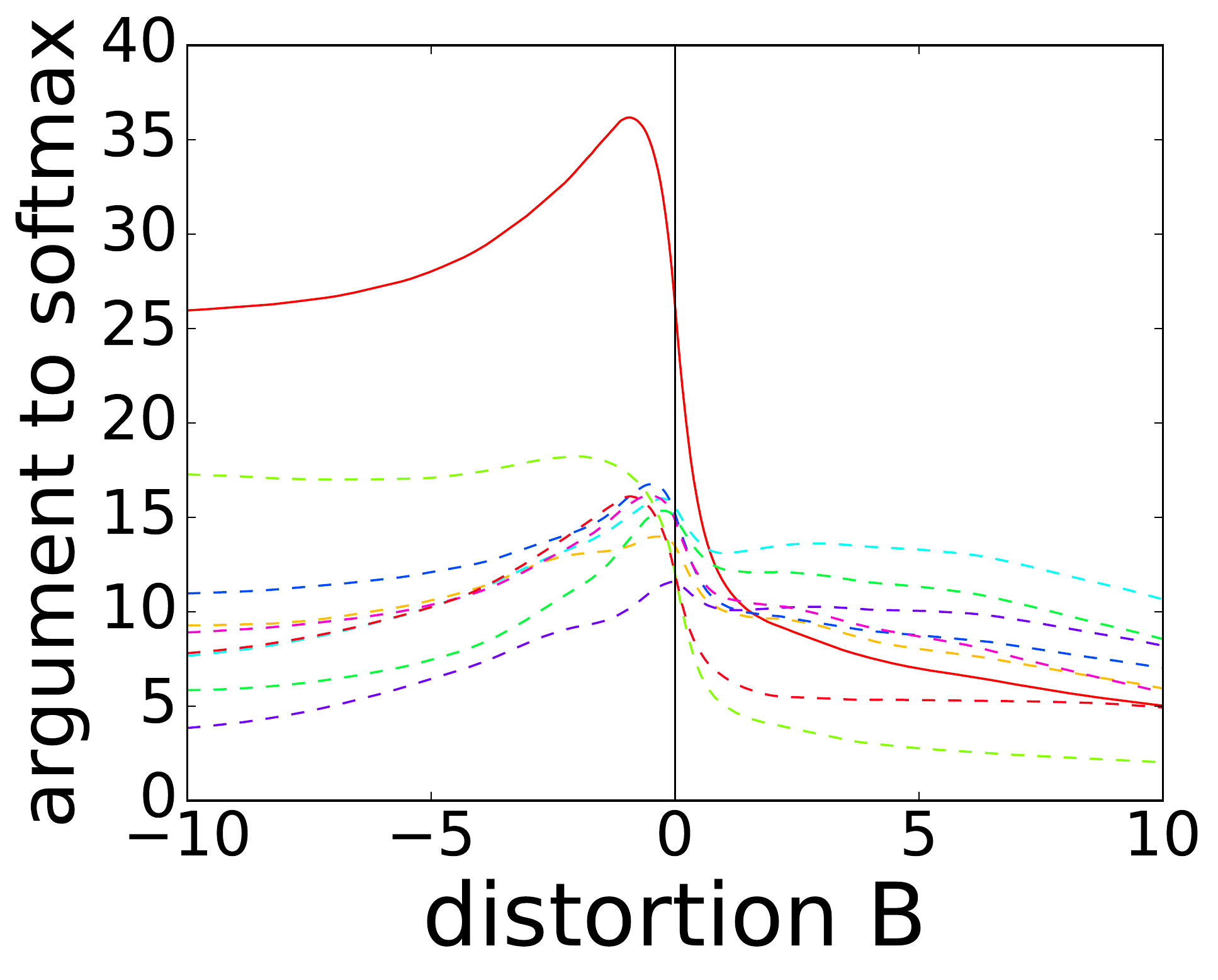}
  \caption{VGG-16}
\end{subfigure}
\begin{subfigure}[b]{0.32\textwidth}
\centering
  \includegraphics[scale=0.2]{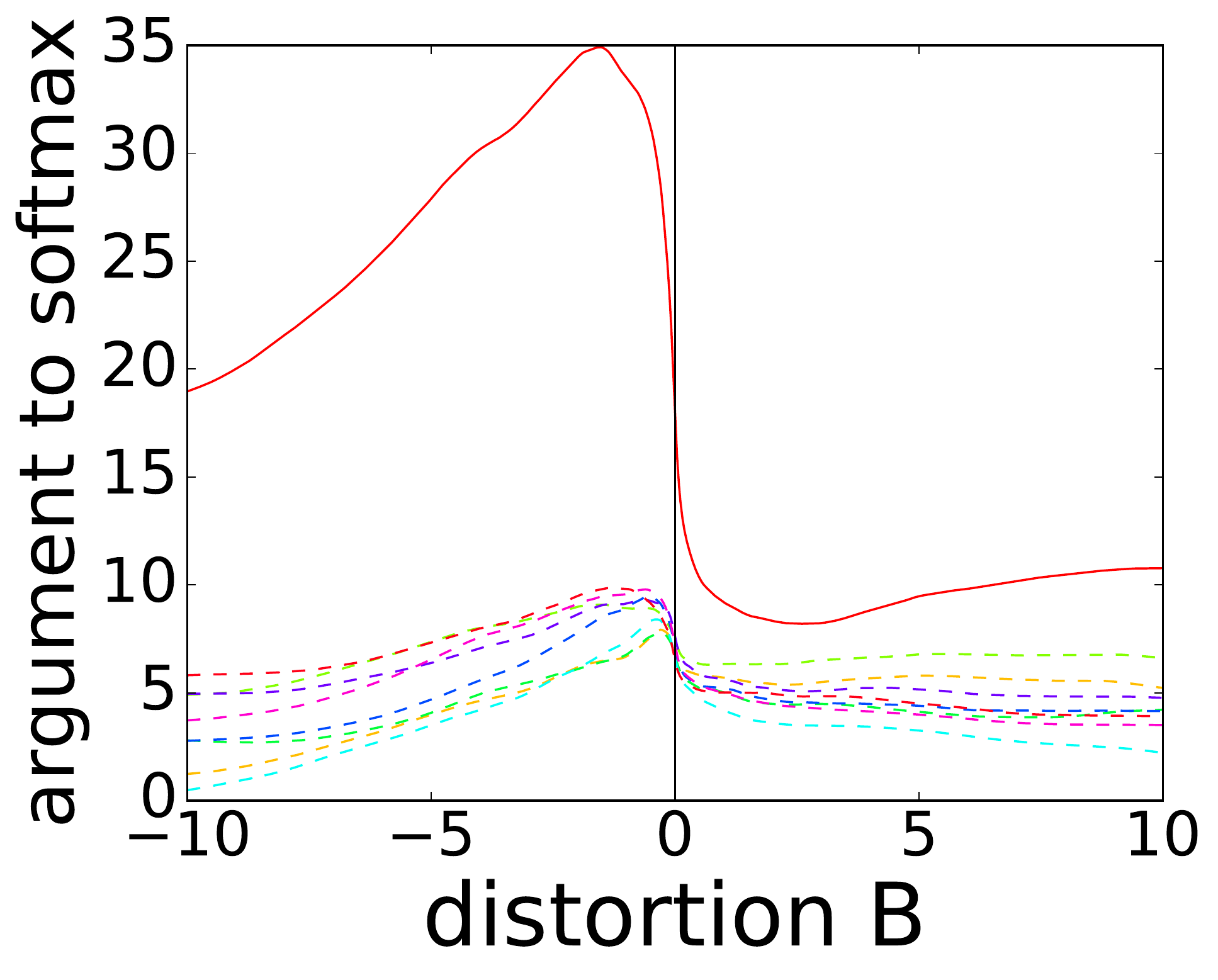}
  \caption{ResNet-152}
\end{subfigure}
\begin{subfigure}[b]{0.49\textwidth}
\centering
  \includegraphics[scale=0.2]{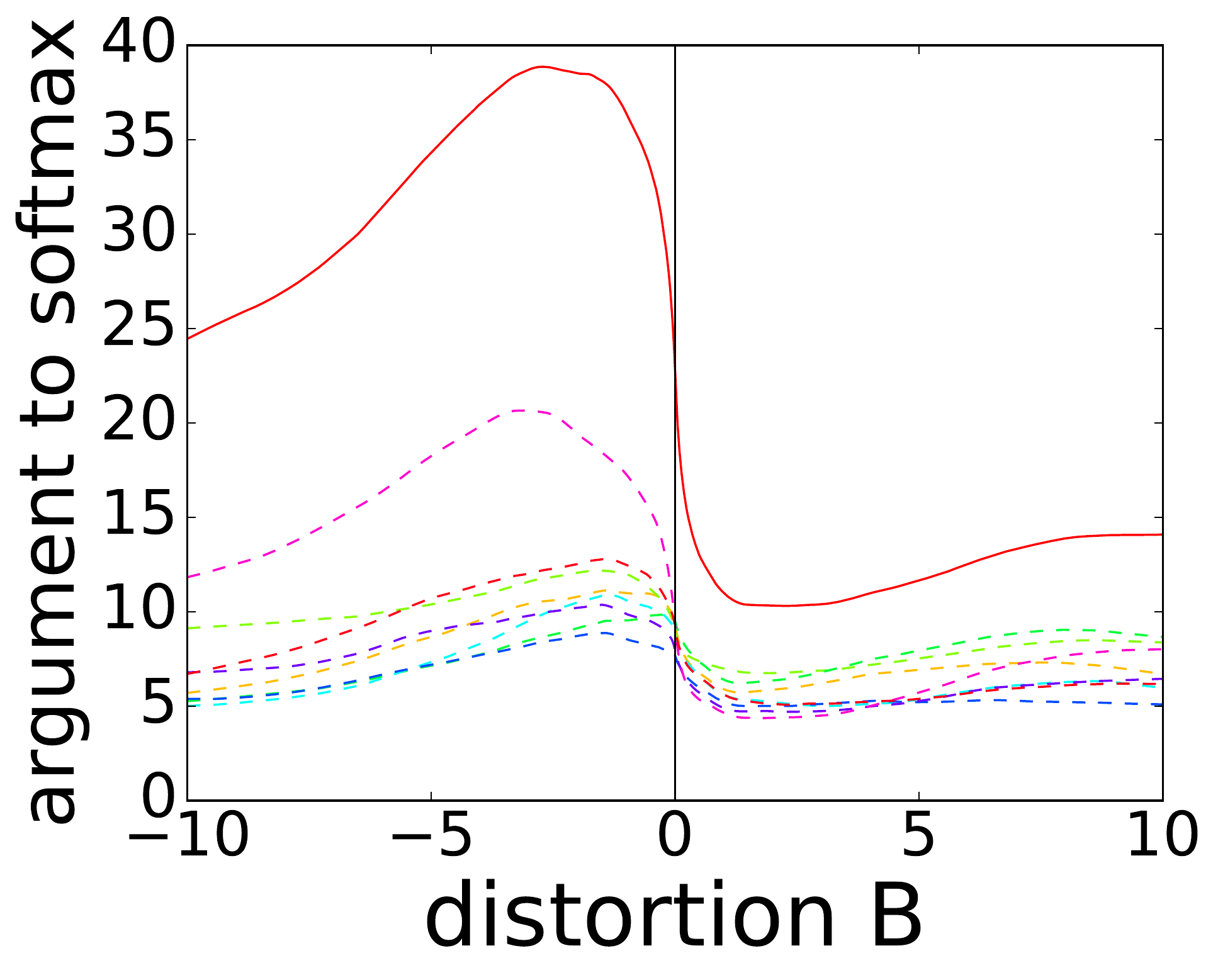}
  \caption{ResNet-101}
\end{subfigure}
\begin{subfigure}[b]{0.49\textwidth}
\centering
  \includegraphics[scale=0.2]{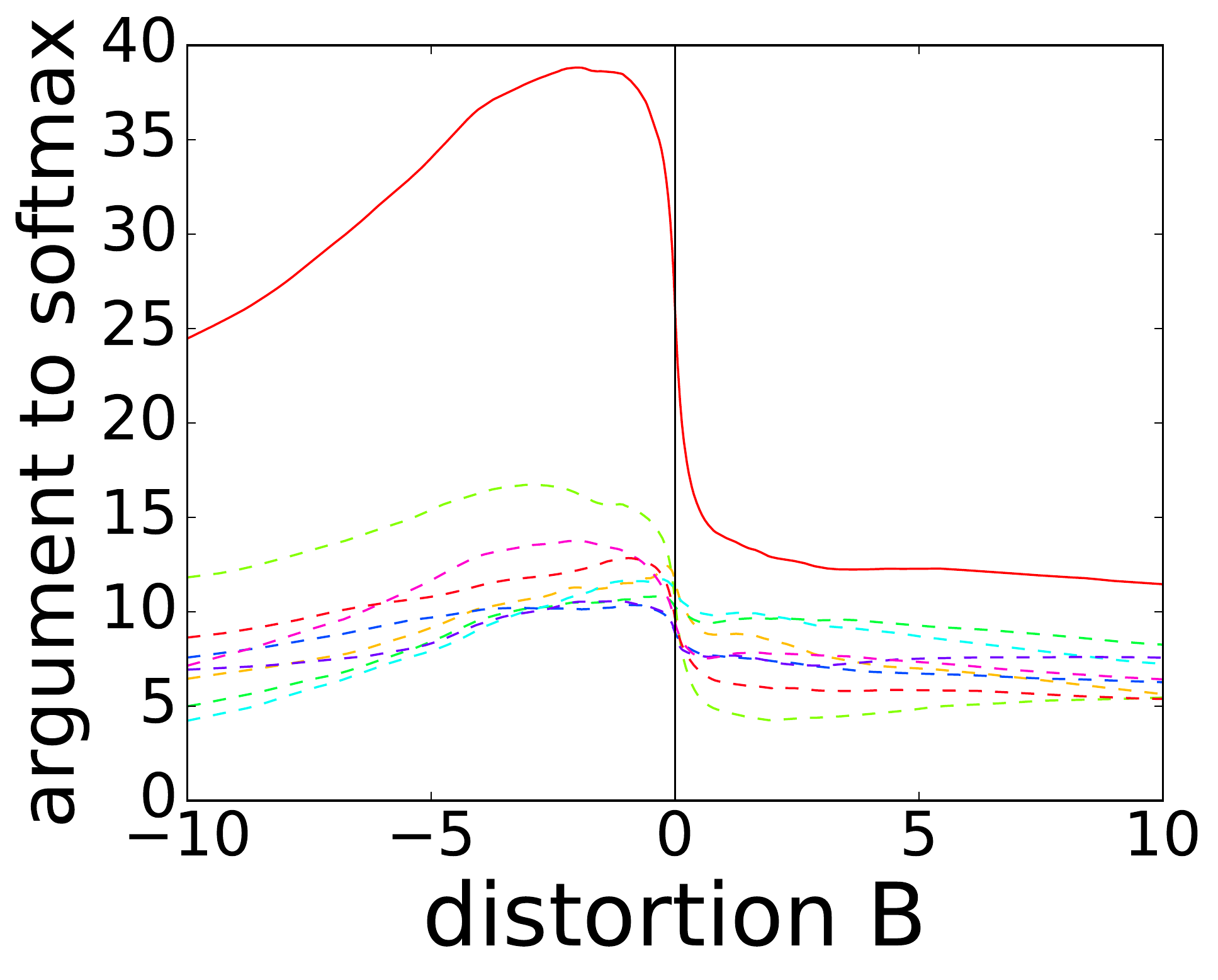}
  \caption{ResNet-50}
\end{subfigure}
\caption{Linearity of different models. Each line plots the classification layer's softmax input vs distortion $B$
when fast gradient noises are used. Solid red line is the ground truth label, and other lines are the top 10 labels
predicted on the original image.}
\label{fig:linearity}
\end{figure}


\paragraph{More examples submitted to \clarifai.} We present more results from \clarifai by submitting original and adversarial examples in Table~\ref{tab:real-app}.

\begin{longtable}{|C{2cm}|C{1.25cm}|C{2.25cm}|C{1.5cm}|C{2cm}|C{3cm}|}\hline
original image & true label & \clarifai results of original image & target label & targeted adv-example & \clarifai result of targeted adversarial example\\ \hline

\includegraphics[width=20mm, height=20mm]{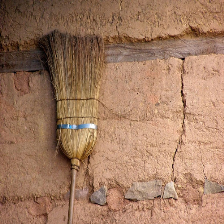} & broom & dust, \newline brick, \newline rustic, \newline stone, \newline dirty & jacamar & \includegraphics[width=20mm, height=20mm]{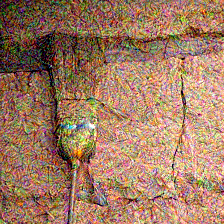} & feather, \newline beautiful, \newline bird, \newline leaf, \newline flora\\ \hline

\includegraphics[width=20mm, height=20mm]{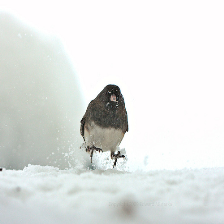} & junco, snowbird & frost, \newline sparrow, \newline pigeon, \newline ice, \newline frosty & eel & \includegraphics[width=20mm, height=20mm]{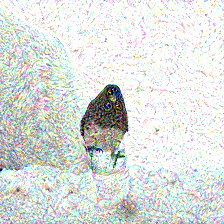} & swimming, \newline underwater, \newline fish, \newline water, \newline one\\ \hline


\includegraphics[width=20mm, height=20mm]{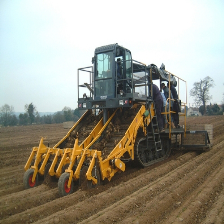} & harvester, reaper & heavy, \newline bulldozer, \newline exert, \newline track, \newline plow & prairie chicken, prairie grouse, prairie fowl & \includegraphics[width=20mm, height=20mm]{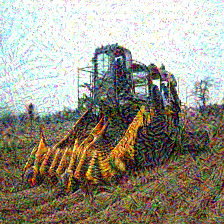} & wildlife, \newline animal, \newline illustration, \newline nature, \newline color\\ \hline

\includegraphics[width=20mm, height=20mm]{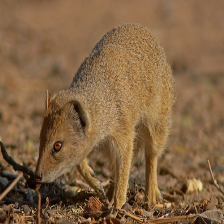} & mongoose & fox, \newline rodent, \newline predator, \newline fur, \newline park & maillot & \includegraphics[width=20mm, height=20mm]{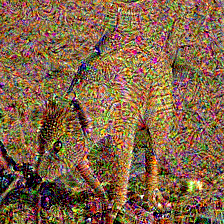} & motley, \newline shape, \newline horizontal, \newline abstract, \newline bright\\ \hline

\includegraphics[width=20mm, height=20mm]{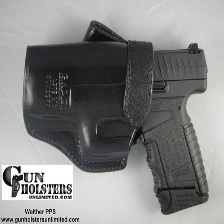} & holster & pistol, \newline force, \newline bullet, \newline protection, \newline cartridge & ground beetle, carabid beetle & \includegraphics[width=20mm, height=20mm]{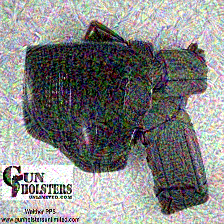} & shell, \newline shell (food), \newline antenna, \newline insect, \newline invertebrate\\ \hline

\includegraphics[width=20mm, height=20mm]{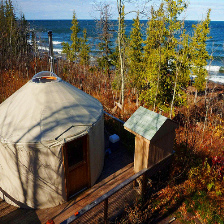} & yurt & wooden, \newline scenic, \newline snow, \newline rural, \newline landscape & comic book & \includegraphics[width=20mm, height=20mm]{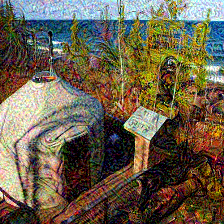} & graffiti, \newline religion, \newline people, \newline painting, \newline culture\\ \hline

\includegraphics[width=20mm, height=20mm]{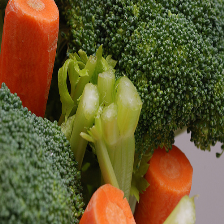} & broccoli & kind, \newline cauliflower, \newline vitamin, \newline carrot, \newline cabbage & hamster & \includegraphics[width=20mm, height=20mm]{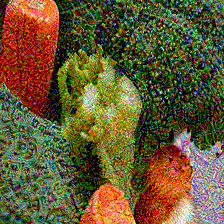} & bird, \newline bright, \newline texture, \newline animal, \newline decoration\\ \hline

\includegraphics[width=20mm, height=20mm]{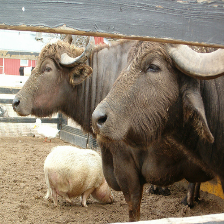} & water buffalo, water ox, Asiatic buffalo, Bubalus bubalis & herd, \newline milk, \newline beef cattle, \newline farmland, \newline cow & rugby ball & \includegraphics[width=20mm, height=20mm]{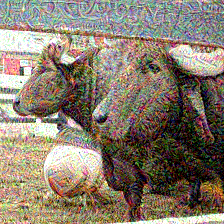} & pastime, \newline print, \newline illustration, \newline art\\ \hline

\includegraphics[width=20mm, height=20mm]{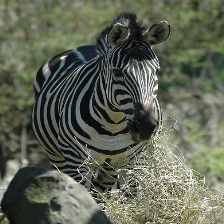} & zebra & equid, \newline stripe, \newline savanna, \newline zebra, \newline safari & apiary, bee house & \includegraphics[width=20mm, height=20mm]{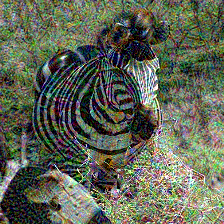} & wood, \newline people, \newline outdoors, \newline nature\\ \hline

\includegraphics[width=20mm, height=20mm]{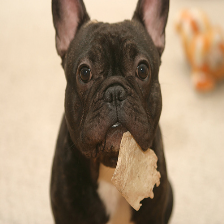} & French bulldog & bulldog, \newline studio, \newline boxer, \newline eye, \newline bull & kite & \includegraphics[width=20mm, height=20mm]{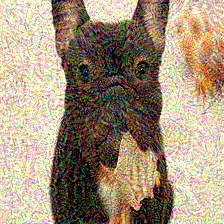} & visuals, \newline feather, \newline wing, \newline pet, \newline print\\ \hline

\includegraphics[width=20mm, height=20mm]{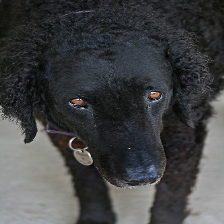} & curly-coated retriever & eye, \newline looking, \newline pet, \newline canine, \newline dog & fly & \includegraphics[width=20mm, height=20mm]{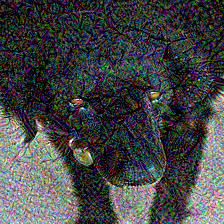} & graphic, \newline shape, \newline insect, \newline artistic, \newline image\\ \hline

\caption{Original images and adversarial images evaluated over \clarifai.
For labels returned from \clarifai, we sort the labels firstly by the occurrence of a label from the \clarifai results, and secondly by confidence. Only top 5 labels are provided.}

\label{tab:real-app}

\end{longtable}

\end{document}